\documentclass[11pt]{article}

\title{Stochastic Parallel Block Coordinate Descent for Large-scale Saddle Point Problems}

\author{Zhanxing Zhu\footnotemark[1]\ \and Amos J. Storkey\footnotemark[1]}

\date{\today}

\usepackage{cite}
\usepackage{url}
\usepackage{subfigure}

\usepackage{amsmath}
\usepackage{graphicx} 
\usepackage{subfigure}
\usepackage{amsthm}
\usepackage{rotating}
\usepackage{multirow}
\usepackage{titlesec}
\usepackage{verbatim}

\usepackage{latexsym}
\usepackage{mathrsfs}
\usepackage{amsmath}
\usepackage{amssymb}
\usepackage{shortbold}

\usepackage{fancyhdr}

\usepackage[margin=3.0cm]{geometry}

\usepackage{enumitem}

\newtheorem{theorem}{Theorem}

\usepackage{algorithm}
\usepackage{algorithmic}
\usepackage{pbox}

\newcommand{\cut}[1]{}

\begin{document}

\maketitle

\renewcommand{\thefootnote}{\fnsymbol{footnote}}

\footnotetext[1]{School of Informatics, University of Edinburgh, EH8 9AB, UK. \texttt{\{zhanxing.zhu, a.storkey\}@ed.ac.uk}.}

\begin{abstract}
  We consider convex-concave saddle point problems with a \emph{separable} structure and \emph{non-strongly convex} functions. We propose an efficient stochastic block coordinate descent method using \emph{adaptive} primal-dual updates, which enables flexible parallel optimization for large-scale problems. Our method shares the efficiency and flexibility of block coordinate descent methods with the simplicity of primal-dual methods and utilizing the structure of the separable convex-concave saddle point problem.  It is capable of solving a wide range of machine learning applications, including robust principal component analysis, Lasso, and feature selection by group Lasso, etc.  Theoretically and empirically, we demonstrate significantly better performance than state-of-the-art methods in all these applications.
\end{abstract}

\pagenumbering{arabic}

\section{Introduction}
A large number of machine learning (ML) models can be cast as convex-concave saddle point (CCSP) problems. There are two common cases. First, convex optimization problems with linear constraints can easily be reformulated as CCSP problems by introducing Lagrangian multipliers~\cite{chen2001,boyd2011,wainwright2014}. Second, empirical risk minimization with regularization (ERM, ~\cite{hastie2009}) can be reformulated as CCSP problem by conjugate dual transformation.  In machine learning applications, these two groups of CCSP problems often exhibit a separable additive structure. Developing efficient optimization methods for seperable CCSP problems is especially important for large-scale applications. Existing work, such as~\cite{zhang2015,zhu2015}, assumes the strong convexity of each of the separable functions, and applies to ERM problems. Although the strong convexity assumption can be relaxed, there is no guide on how to select the extra regularization parameters. We also find the relaxation significantly hinders convergence rates even for post-hoc optimal choices of parameters\cut{; the regularization overly constrains the size of steps that are possible}. Furthermore, inappropriate parameter selection dramatically deteriorates the practical performance. Even for strongly-convex systems the strong-convexity parameter is often hard to determine. Additionally, it is currently unclear how to adapt the stepsize for handling block separable problems.

In this work, we propose a novel stochastic and parallelizable approach for Sep-CCSP problem, which naturally handles convex cases that are not strongly convex and avoids any notorious hyperparameter selection issues. This method is also capable of dealing with block separable CCSP problem. In the following, we formally introduce the Sep-CSSP problem and consider the two common machine learning instantiations of this problem.

The generic \textbf{convex-concave saddle point problem} is written as
\begin{equation}
\min_{\xB \in \Rbb^n} \max_{\yB \in \Rbb^{m}} \left\{ L(\xB, \yB) = f(\xB) + \langle \yB, \AB \xB \rangle - g^{*}(\yB) \right\}, \label{eq:origin}
\end{equation}
where $f(\xB)$ is a proper convex function, $g^*$ is the convex conjugate of a convex function $g$, and $\AB \in \Rbb^{m \times n}$. Many machine learning tasks reduce to solving a problem of this form.
One important subclass of~(\ref{eq:origin}) is where $f(\xB)$ or $g^*(\yB)$ exhibits an additive {separable structure}. We say $f(\xB)$ is \emph{separable} when $f(\xB) = \sum_{j=1}^J f_j(\xB_j)$, with $\xB_j \in \Rbb^{n_j}$, and $\sum_{j=1}^J n_j = n$. Separability for $g^*(\cdot)$ is defined likewise. We can also partition matrix $\AB$ into $J$ column blocks $\AB_j \in \Rbb^{m \times n_j}$, $j = 1,\dots,J$,  and $\AB \xB = \sum_{j=1}^J  \AB_j \xB_j$, resulting in a problem of the form
\begin{equation}
\min_{\xB \in \Rbb^n} \max_{\yB \in \Rbb^{m}}   \sum_{j=1}^J f_j(\xB_j) +  \sum_{j=1}^J \langle \yB, \AB_j \xB_j \rangle - g^{*}(\yB). \label{eq:sepccspf}
\end{equation}
We call problems of the form (\ref{eq:sepccspf}) \textbf{Separable Convex Concave Saddle point (\emph{Sep-CCSP})} problems. We develop an efficient optimization method for Sep-CCSP problems when $f(\cdot)$ and/or $g^*(\cdot)$ are \emph{non-strongly convex}; many ML methods result in a non-strongly convex Sep-CCSP form.

\paragraph{Example 1} Separable function minimization with linear constraints takes the form
\begin{align}
\min_{\xB} \sum_{i=1}^J & f_i(\xB_i) \mbox{ s.t. }  \sum_{i=1}^{J} \AB_i \xB_i = \bB, 
\label{eq:linear_constrained}
\end{align}
leading to
\begin{equation}
\min_{\xB} \max_{\yB} L(\xB, \yB) =  \sum_{i=1}^J f_i(\xB_i) + \langle \yB, \sum_{i=1}^J  \AB_i \xB_i \rangle - \yB^T \bB
\end{equation}
when we introduce Lagrangian multipliers $\yB$ for the linear constraints. Here $g^*(\yB) = \yB^T \bB$ is non-strongly convex.
A large number of machine learning problems can be expressed as linearly constrained optimization problems of this form~\cite{chen2001,boyd2011,wainwright2014}, for instance, robust principal component analysis (RPCA)~\cite{wright2009,candes2011}. 

\paragraph{Example 2} Another important case of Sep-CCSP is empirical risk minimization (ERM, \cite{hastie2009}) of linear predictors, with a convex regularization function $f(\xB)$:
\begin{equation}
\min_{x} P(\xB) =  f(\xB) + \frac{1}{N} \sum_{i = 1}^{N} g_i(\aB_i^T \xB)
\end{equation}
where $N$ labels the number of data points. Many well-known classification and regression problems are included in this formulation, such as group Lasso \cite{yuan2006} with the regularizer as a sum of groupwise $L_2$-norm $f(\xB) = \sum_{g=1}^G f_g(\xB_g) = \lambda \sum_{g=1}^G w_g \| \xB_g \|_2$. Reformulating the above regularized ERM by employing the conjugate dual of function $g$, i.e.,
\begin{equation}
g_i(\aB_i^T \xB) = \sup_{y_i \in \Rbb} y_i\langle \aB_i, \xB \rangle - g_i^*(y_i), \label{eq:conjugatedual}
\end{equation}
we transform it into a Sep-CCSP problem,
\begin{equation}
\min_{\xB} \max_{\yB} \sum_{g=1}^G f_g(\xB_g) + \frac{1}{N} \langle \sum_{i=1}^N y_i \aB_i,  \xB \rangle - \frac{1}{N}\sum_{i=1}^N g_i^*(y_i). \label{eq:erm_convex_concave}
\end{equation}
If $g_i(\cdot)$ is not smooth (e.g. hinge or absolute loss), the conjugate dual $g^*(\cdot)$ is non-strongly convex.



Inspired by current active research on block coordinate descent methods (BCD, \cite{nesterov2012efficiency,richtarik2015parallel,richtarik2014iteration}), we propose a Stochastic Parallel Block Coordinate Descent method (SP-BCD) for solving the separable convex-concave saddle point problems, particularly non-strongly convex functions. The key idea is to apply stochastic block coordinate descent of the separable primal space into the primal-dual framework~\cite{chambolle2014,pock2011} for the Sep-CCSP problem. We propose a novel \emph{adaptive} stepsize for both the primal and dual updates to improve algorithm convergence performance.  Compared with the standard primal-dual framework, our method enables the selected blocks of variables to be optimized in parallel according to the processing cores available. Without any assumption of strong convexity or smoothness, our method can achieve an $O(1/T)$ convergence rate, which is the best known rate for non-strongly (and non-smooth) convex problem. Also, in a wide range of applications, we show that SP-BCD can achieve significantly better performance than the aforementioned state-of-the-art methods. These results are presented in Section \ref{sec:app}.

The authors in \cite{wang2014} proposed a stochastic and parallel algorithm for solving the problem~(\ref{eq:linear_constrained}). However, their method is based on an augmented Lagrangian, often suffering from the selection of penalty parameter.  As previously discussed, the methods for handling Sep-CCSP in~\cite{zhang2015,zhu2015} 
focused on the ERM problem, and assumed that both $f(\xB)$ and $g^{*}(\yB)$ are strongly convex, or relaxed that constraint in ways that we show significantly hits performance, and required additional hyperparameter selection (as do augmented Lagrangian methods). Additionally, the method in~\cite{zhang2015} is not capable of handling block separable CCSP problem. These all limit its applicability.  Our approach SP-BCD can overcome these difficulties, which can (i) naturally handle non-strongly convex functions, and avoids any notorious hyperparameter selection issues; (ii) is capable of handling block separable CCSP problem. 


\section{Primal-dual Framework for CCSP}
\label{sec:pd}
In \cite{chambolle2011}, the authors proposed a first-order primal-dual method for (non-smooth) convex problems with saddle-point structure, i.e., Problem (\ref{eq:origin}). We refer this algorithm as PDCP. The update of PDCP in $(t+1)$-th iteration is  as follows:
\begin{align}
\yB^{t+1} &= \argmin_{\yB} g^{*}(\yB) - \langle \yB,  \AB \overline{\xB}^t \rangle + \frac{\sigma}{2} \| \yB - \yB^t  \|_2^2 \\
\xB^{t+1} &= \argmin_{\xB} f(\xB) + \langle \yB^{t+1}, \AB \xB\rangle + \frac{h}{2} \| \xB - \xB^t  \|_2^2 \\
\overline{\xB}^{t+1} &= \xB^{t+1} + \theta (\xB^{t+1} - \xB^t). \label{eq:extra_origin}
\end{align}
When the parameter configuration satisfies $\sigma h \geq \| \AB \|^2$ and $\theta = 1$, PDCP can achieve a $O(1/T)$ convergence rate. For the general CCSP problem, PDCP does not consider the structure of matrix $\AB$ and only applies constant stepsize for all dimensions of primal and dual variables. Based on PDCP, \cite{pock2011} used the structure of matrix $\AB$ and proposed a diagonal preconditioning technique for PDCP, which showed better performance in several computer vision applications.
However, when the function $f(\xB)$ has separable structure with many blocks of coordinates, both these algorithms are batch methods and non-stochastic, i.e. they have to update all the primal coordinates in each iteration. This influences empirical efficiency.

Inspired by the recent success of coordinate descent methods for solving separable optimization problems, we incorporate a stochastic block coordinate descent technique into above primal-dual methods and propose adaptive stepsizes for the chosen blocks via the structure of the matrix $\AB$. 


\section{Our Method: SP-BCD for Sep-CCSP}
\label{sec:spbcd}
The basic idea of our stochastic parallel block coordinate descent (SP-BCD) method for solving the saddle point problem (\ref{eq:sepccspf}) is simple; we optimize $L(\xB, \yB)$ by alternatively updating the primal and dual variables in a principled way. Thanks to the separable structure of $f(\xB)$, in each iteration we can randomly select $K$ blocks of variables whose indices are denoted as $S_t$, and then we only update these selected blocks, given the current $y=y^t$, in the following way. If $ j \in S_t$ then
\begin{equation}
\xB_j^{t+1} =\argmin_{\xB_j} f_j(\xB_j) + \langle \yB^t, \AB_j \xB_j \rangle + \frac{1}{2} \| \xB_j - \xB_j^t \|_{\hB_j}^2,
\label{eq:primalupdate}
\end{equation}
otherwise, we just keep $\xB_j^{t+1} = \xB_j^t$.
In the blockwise update, we add a proximal term to penalize the deviation from last update $\xB^t_j$, i.e.,
\begin{equation}
\frac{1}{2} \| \xB_j - \xB_j^t \|_{\hB_j}^2 = \frac{1}{2} (\xB_j - \xB_j^t)^T \diag (\hB_j) (\xB_j - \xB_j^t),
\end{equation}
where the diagonal matrix $\HB_j = \diag (\hB_j) $ is applied for scaling each dimension of $\xB_j$, and each $\hB_j$ is a subvector of $\hB = \left[ \hB_1^T, \dots, \hB_J^T \right]^T$. We configure the each dimension of $\hB$ as
\begin{equation}
\quad h_{d} = \sum_{j=1}^m |  A_{jd} |, \quad d = 1, 2, \dots, n. \label{eq:primal_stepsize}
\end{equation}
Intuitively, $h_d$ in our method can be interpreted as the coupling strength between the $d$-th dimension of the primal variable $\xB$ and dual
variable $\yB$, measured by the $L_1$ norm of the vector $\AB_{:,d}$ (i.e., the $d$-th column of matrix $\AB$). Smaller coupling strength
allows us to use smaller proximal penalty (i.e., larger stepsize) for updating the current primal variable block without caring
too much about its influence on dual variable, and vice versa.

Then for those selected block variables, we use an extrapolation technique given in Eq.(\ref{eq:extra_origin}) to yield an intermediate variable $\overline{\xB}^{t+1}$ as follows,
 \begin{equation}
	\overline{\xB}_j^{t+1} =
	\begin{cases}
	\xB_j^{t+1} + \theta \left( \xB_j^{t+1} - \xB_j^t \right)  & \text{if } j \in S_t\\
	\overline{\xB}_j^t & \text{otherwise},
	\end{cases}
	\label{eq:extrapolation}
\end{equation}
where $\theta = K/J$ to account for there being only $K$ blocks out of $J$ selected in each iteration.

Assuming $g^*(\yB)$ is not separable, we update the dual variable as a whole. A similar proximal term is added with the diagonal matrix $\SigmaB^t = \diag ( \sigmaB^t)$:
\begin{multline}
 \yB^{t+1} = \argmin_{\yB}  g^*(\yB) - \langle \yB, \overline{\rB}^t + \frac{J}{K} \sum_{j \in S_t} \AB_j (\overline{\xB}_j^{t+1} - \overline{\xB}_j^{t} ) \rangle \\ 
                                     + \frac{1}{2} \| \yB - \yB^t \|^2_{\sigmaB^t}, \label{eq:dualupdate}
\end{multline}
where $\overline{\rB}^{t} = \sum_{j=1}^J \AB_j \overline{\xB}_j^{t} $. We configure the dual proximal penalty $\sigmaB^t$ \emph{adaptively} for each iteration,
\begin{equation}
\sigma_k^t = \frac{J}{K} \sum_{j \in S_t} |A_{kj} |, \quad k = 1,2,\dots,m. \label{eq:dual_stepsize}
\end{equation}
This configuration adaptively accounts for the coupling strength between the dual variable and the chosen primal variable blocks in $S_t$ through measuring the structure of the matrix $\AB$. Later we show that the usage of the proposed adaptive proximal penalty for both primal and dual update contributes to significantly improve the convergence performance for many machine learning applications.

Another crucial component of the dual update is the construction of the term $ \overline{\rB}^t + \frac{J}{K} \sum_{j \in S_t} \AB_j (\overline{\xB}_j^{t+1} - \overline{\xB}_j^{t} )$, which is inspired by a recently proposed fast incremental gradient method for non-strongly convex functions, SAGA~\cite{defazio2014saga}. We use the combination of the cached sum of all $\AB_j \overline{\xB}_j^t$, i.e., $\overline{\rB}^t$, and the newly updated sample average $\frac{1}{K} \sum_{j \in S_t} \AB_j (\overline{\xB}_j^{t+1} - \overline{\xB}_j^{t} )$ to obtain a variance reduced estimation of $\Ebb [\overline{\rB}]$, which is essentially the spirit of SAGA.
After the dual update, $\overline{\rB}^t$ is updated to $\overline{\rB}^{t+1}$ using,
   \begin{equation}
   \overline{\rB}^{t+1} = \overline{\rB}^t + \sum_{j \in S_t}  \AB_j \left( \overline{\xB}^{t+1}_j -  \overline{\xB}^{t}_j \right).
   \label{eq:rupdate}
   \end{equation}

\begin{algorithm}[tb]
   \caption{SP-BCD for Separable Convex-Concave Saddle Point Problems}
   \label{alg:SP-BCD}
\begin{algorithmic}[1]
   \STATE {\bfseries Input:} number of blocks picked in each iteration $K$, $\theta= K/J$, the configuration of $\hB$ and $\sigmaB^t$ as given in Eq.~(\ref{eq:primal_stepsize}) and (\ref{eq:dual_stepsize}), respectively.
   \STATE {\bfseries Initialize:} $\xB^0$, $\yB^0$, $\overline{\xB}^0 = \xB^0$, $\overline{\rB}^0 = \sum_{j=1}^J \AB_j \overline{\xB}_j^{0}$
   \FOR{$t=1,2,\ldots, T$}
   \STATE Randomly pick set $S_t$ of $K$ blocks from $\{ 1, \dots, J \}$ each chosen with probability $K/J$.
   \FOR{each block in parallel}
   \STATE Update each primal variable block using Eq.(\ref{eq:primalupdate}), and extrapolate it using Eq.(\ref{eq:extrapolation});
   \ENDFOR
   \STATE Update dual variables using Eq.(\ref{eq:dualupdate}) and update $\overline{\rB}^{t+1}$ using Eq.~(\ref{eq:rupdate}).
   \ENDFOR
\end{algorithmic}
\end{algorithm}
The whole procedure for solving Sep-CCSP problem (\ref{eq:sepccspf}) using SP-BCD is summarized in Algorithm \ref{alg:SP-BCD}. There are several notable characteristics of our algorithm:

\emph{1. }This algorithm is amenable to parallelism for large-scale optimization, which is suitable for modern computing clusters. Our method possesses one of key advantages of stochastic parallel coordinate descent method~\cite{richtarik2015parallel}: providing the flexibility that in each iteration the number of selected blocks can be optimized completely in parallel according to available number of machines or computational cores. This could make use of all the computational availability as effectively as possible.

\emph{2. }The related non-stochastic primal-dual algorithms~\cite{chambolle2011,chambolle2014} need evaluation of the norm of $\AB$. For large problem size, the norm evaluation can be time-consuming. The parameter configuration in our algorithm avoids norm estimation, but maintains a $O(1/T)$ convergence rate.

%

\emph{3. }Compared with recent work \cite{zhang2015}, we do not assume strong convexity of $f(\xB)$ and $g^*(\yB)$. This removes the need to regularise and improves applicability, as demonstrated in Section \ref{sec:app}.

\emph{3. } Although an augmented Lagrangian framework, such as ADMM, can implement an effective optimization for many problems with linear constraints~(\ref{eq:linear_constrained}), the selection of the penalty parameter has a dramatic influence on its performance. Current selection rules rely on various heuristics or exhaustive search, and no theoretical justifications exist. This difficulty also occurs with other recent work ~\cite{zhang2015} when $f(\xB)$ and $g^*(\yB)$ are not strongly convex. Our method avoids this issue.

\subsection*{Convergence Analysis}
For a convergence analysis, we employ the following gap for any saddle point $(\xB, \yB)$, 
$\Gcal (\xB', \yB') \triangleq \max_{\yB} L(\xB', \yB) - \min_{\xB} L(\xB, \yB' ).$
As discussed by \cite{chambolle2011}, this gap will practically measure the optimality of the algorithm if the domain of the $(\xB', \yB')$ is `` large enough" such that  $(\xB', \yB')$ could lie in the interior of their domains. The following theorem establishes the convergence of our algorithm.

\begin{theorem}
Given that all $f_i(\xB_i)$ and $g^*(\yB)$ are convex functions,  and we set $\theta = K/J$, proximal parameters for primal and dual update as Eq.(\ref{eq:primal_stepsize}) and (\ref{eq:dual_stepsize}), respectively.
Then for any saddle point $(\xB, \yB)$, the expected gap decays as the following rate:
\begin{equation*}
\Ebb \left[ L\left(\sum_{t=1}^T \xB^t/T, \yB \right) - L\left(\xB, \sum_{t=1}^T \yB^t/T \right)  \right]  \leq  \frac{1}{T}  M(0),
\end{equation*}
where $M(0) =$
\begin{align*}
& \frac{J}{2K} \| \xB^0 - \xB \|_{\hB}^2 + \frac{1}{2} \|  \yB^0 - \yB  \|_{\sigmaB^0}^2
 - \langle  \yB^0 - \yB, \AB \left( \xB^0 - \xB \right) \rangle \\ 
&  
  +\frac{J-K}{K} \left( f(\xB^0) + \langle \yB, \AB \xB^0 \rangle - \left( f(\xB) +   \langle \yB, \AB \xB  \rangle  \right)    \right).
\end{align*}
\label{th:main}
\end{theorem}
\vskip -0.5cm
The proof of the above theorem is technical and given in Appendix.

\emph{Remark.} For the parameter configuration in Theorem \ref{th:main}, when $\theta = K/J$, the key point for obtaining the convergence of our algorithm is that we select one particular configuration of $\hB$ and $\sigmaB^t$ to guarantee the positive semidefiniteness of the following matrix,
\begin{equation}
\PB =
\begin{bmatrix}
\diag (\hB_{S_t}) & -\AB_{S_t}^T \\
-\AB_{S_t} & \frac{K}{J} \diag (\sigmaB^t)
\end{bmatrix}
\succeq 0.
\end{equation}
Under the parameter configuration of $\hB$ and $\sigmaB^t$ in Theorem \ref{th:main}, we can guarantee matrix $\PB$ is diagonally dominant, directly leading positive semidefiniteness. However, the parameter configuration to make $\PB \succeq 0$ is not unique. We find that other configurations are also valid, for instance, for each block $j$, $\hB_j = \| \AB_j \| \IB$ and $\sigmaB = \frac{J}{K} \sigma \IB$, where $\sigma = \max \{ \| \AB_j \| \}_{j=1}^J$. Different parameter configuration might provide some influence on the performance of the algorithm. We leave the comparison between them and further theoretical analysis as future work.

%

\section{Applications}
\label{sec:app}
In this section, we provide examples of Sep-CCSP problems in machine learning. In each application, we select different methods to compare with that have already shown strong performance in that particular scenario. Note that,  since the method in~\cite{zhang2015} cannot handle block separable CCSP problem,  it is not applicable for the first and third experiment. To provide a fair comparison with other methods, all the experiments are implemented in one core/machine.  Each experiment is run 10 times and the average results are reported to show statistical consistency.
\begin{table}[ht!]
\caption{\small{RPCA problem: performance of all compared methods (with ADMM, GSADMM and PDMM hyperparameters set to the post-hoc optimal). 
}
}
\centering
\small{
\begin{tabular}{c|c|c|c|c}
\hline
Methods    & Iteration  & Time (s)  &\pbox{20cm}{Frobenus norm\\ of residual ($10^{-4}$) }  &   Objective ($10^8$) \\
\hline
ADMM       & 149       & 2191      &            9.71                       &  1.924 \\
\hline
GSADMM     & $\bold{23}$         & $\bold{448}$       &            8.69                       &  1.924 \\
\hline
PDCP       & 59         & 911       &            7.80                       &  1.924 \\
\hline
PDMM1      & 125        & 927       &            9.92                       &  1.924 \\
PDMM2      & 73         & 750       &            $\bold{4.55}$                       &  1.924 \\
PDMM3      & 67         & 834       &            8.56                       &  1.924 \\
\hline
SP-BCD1    & 104        & 784       &            7.63                       &  1.924 \\
SP-BCD2    & 48         & $\bold{492}$       &  $\bold{6.17}$                       &  1.924 \\
SP-BCD3    & 42         & 553       &            6.72                       &  1.924 \\
\hline
\end{tabular}}
\label{tab:rpca}
\end{table}

\begin{figure}[h!]
\begin{center}
\begin{tabular}{ccc}
\includegraphics[width=0.7\columnwidth]{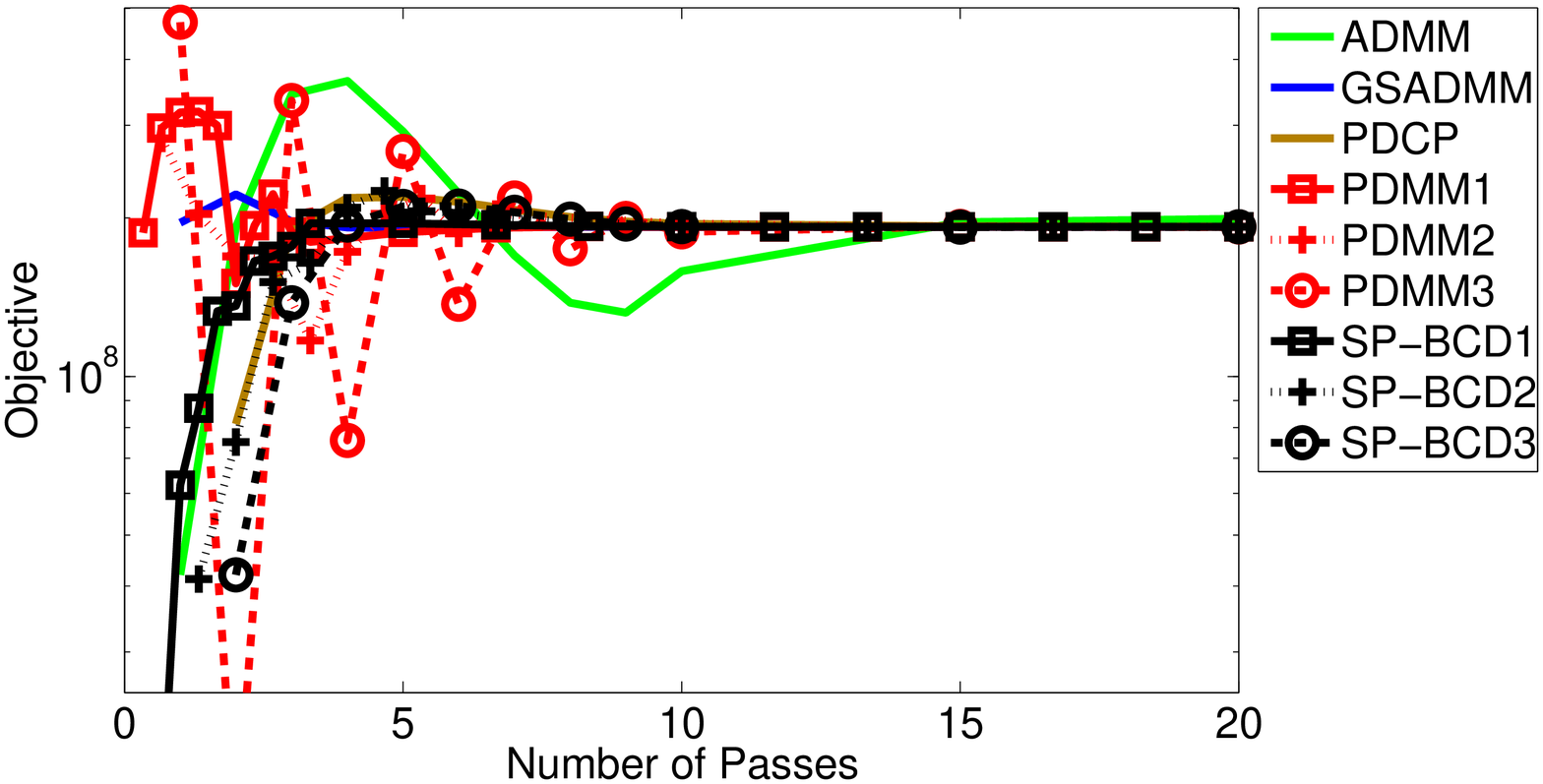}\\
\includegraphics[width=0.4\columnwidth]{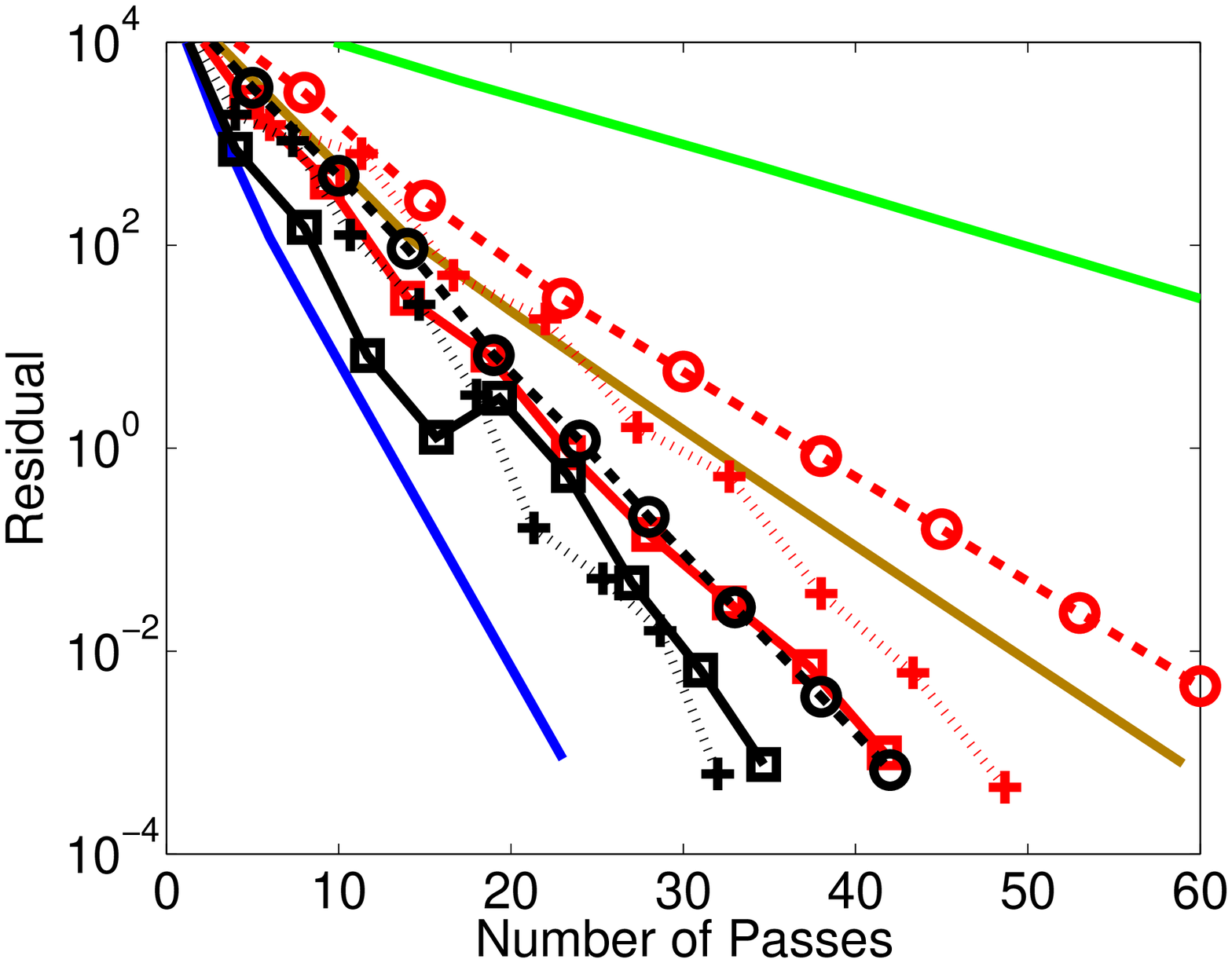} \includegraphics[width=0.4\columnwidth]{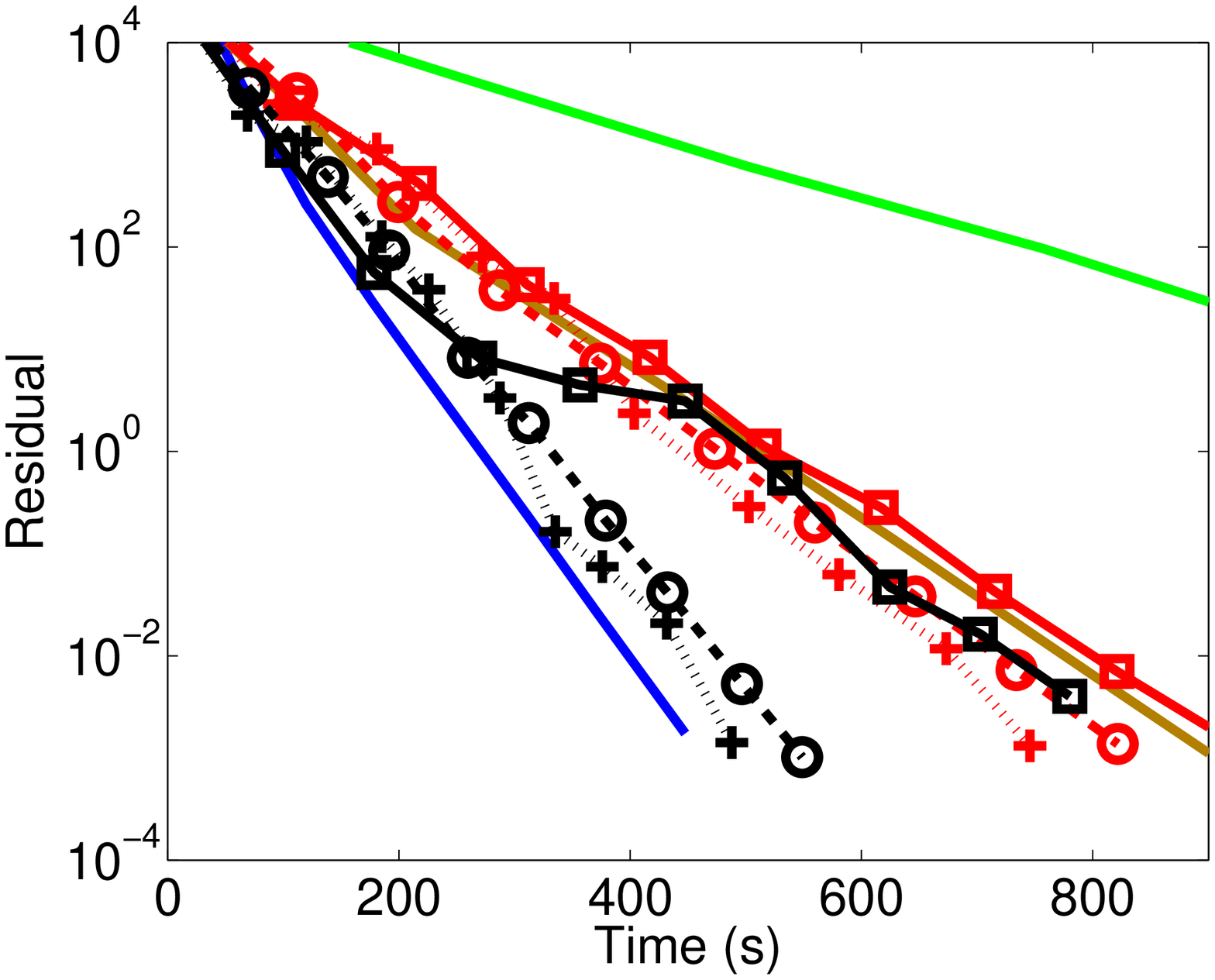}
\end{tabular}
\vspace{-2mm}
\end{center}
\caption{\small RPCA problem: our method (with $K=\{1, 2, 3\}$) versus ADMM, GSADMM, PDCP and PDMM (with $K=\{1, 2, 3\}$). 
\label{fig:rpca}}
\vspace{-2mm}
\end{figure}

\subsection{Robust Principal Component Analysis}
Robust Principal Component Analysis (RPCA) is a variant of PCA to obtain a low rank and sparse decomposition of an observed data matrix $\BB$ corrupted by noise \cite{wright2009,candes2011}, which could help to handle outliers existing in datasets. RPCA aims to solve the following optimization problem,
\begin{equation}
\min_{\{ \XB_i \}_{i=1}^3}  \frac{1}{2} \| \XB_1 \|_{\text{F}}^2 + \mu_2 \| \XB_2\|_1  + \mu_3 \| \XB_3 \|_{*} \quad \text{s.t. } \BB = \sum_{i=1}^3 \XB_i ,  \label{eq:rpca_obj} \nonumber
\end{equation}
where $\BB \in \Rbb^{m \times n}$, $\XB_1$ is a noise matrix, $\XB_2$ is a sparse matrix, $\XB_3$ is a low rank matrix, and $\| \cdot \|_*$ is the nuclear norm of a matrix.
We generate the observation matrix $\BB$ in the same way as \cite{parikh2014}\cut{\footnote{\url{http://stanford.edu/~boyd/papers/prox_algs/matrix_decomp.html}}}, where we have $m=2000$, $n=5000$ and the rank is $r=100$. The regularization parameters are set as $\mu_2 = 0.15 \| \BB \|_{\infty}$ and $\mu_3 = 0.15 \| \BB \|$. Note that RPCA problem with this matrix size is non-trivial since there are in total $30,000,000$ variables and $10,000,000$ equality constraints to handle.

In this particular application, the parameter configuration for SP-BCD with each different number of blocks $K$ chosen from the possible $3$ in each iteration can be obtained: (1) $K=1$, $\left( \theta, h, \sigma^t \right) = \left( 1/3,1,1 \right) $; (2) $K=2$, $\left( \theta, h, \sigma^t \right) = \left( 2/3,1,2 \right)$; (3) $K=3$, $\left(\theta, h, \sigma^t \right) = \left(1, 1,3 \right)$.

Our method SP-BCD is compared with (1) ADMM implemented by~\cite{parikh2013}; (2) Gauss-Seidel ADMM (GSADMM, \cite{hong2012}), which solves the problem (\ref{eq:linear_constrained}) in a cyclic block coordinate manner. However, GSADMM with multiple blocks is not well understood and there is no theory guarantee, and GSADMM has to be implemented sequentially and cannot be parallel; (3) PDCP~\cite{chambolle2011}, for which the recommended parameter configuration can be computed as $\left(\theta, h, \sigma\right) = \left(1, \sqrt{3}, \sqrt{3}\right)$; (4) PDMM \cite{wang2014} with $K=\{1, 2, 3\}$. For each of the three competing methods (ADMM, GSADMM and PDMM)  we run extensive experiments using different penalty parameter  values $\rho$, and report the results for best performing $\rho$, despite the fact that knowledge of which $\rho$ is optimal is not available to the algorithms a priori. Hence the real-world performance of SP-BCD relative to these methods is significantly greater than these figures suggest.

Figure \ref{fig:rpca} depicts the performance of all the methods on the evolution of the objective and the residual (i.e., the deviation from satisfied constraints measured by $\| \XB_1 +  \XB_2 + \XB_3 - \BB \|_{\text{fro}}$) w.r.t. number of passes and consumed time. All methods quickly achieve the consensus objective value in $20$ passes. The key difference in performance is how fast they satisfy the equality constraint. Our method SP-BCD with $K=2$ is the fastest, achieving almost the same performance with GASDMM, while being fully parallelizable whereas GSADMM can only be run sequentially. Although PDMM2 obtains the lowest residual (measured by Frobenus Norm of deviation of satisfied constraints), it spends much longer time $750s$, compared with $492s$ for SP-BCD2. When we run the SP-BCD2 with the same amount of time as that of PDMM2, SP-BCD2 could achieve Frobenus Norm of residual as $\bold{2.36 \times 10^{-4}}$, which shows better performance than PDMM2. The real difference in performance is greater as optimal hyperparameters are not actually available to the competing methods.

\begin{table}[b!]
\vspace{-6mm}
\caption{\small{Lasso problem: performance of all compared methods. 
}}
\centering
\small{
\begin{tabular}{c|c|c|c|c}
\hline
Methods               & $m,n, d$           & Time (s)  &  Number of passes  &  Objective \\
\hline
\multirow{2}{*}{ISTA} & $1000,5000,500$    & 2.27      & 100                                              & 111.405  \\ \cline{2-5}
                      & $5000,20000,2000$   & 45.67     & 100                                              & 448.351 \\
\hline
\multirow{2}{*}{FISTA} & $1000,5000,500$   & 1.16      & 56                                                 & 111.320  \\ \cline{2-5}
                       & $5000,20000,2000$  & 19.00     & 49                 & 448.271 \\
\hline
\multirow{2}{*}{ADMM} & $1000,5000,500$    & $\bold{0.69}$      & 63                   & $\bold{111.318}$  \\ \cline{2-5}
                      & $5000,20000,52000$   & 19.83     & 51                & $\bold{448.258}$\\
\hline
\multirow{2}{*}{PDCP} & $1000,5000,500$    & 1.40      & 100                                     & $\bold{111.318}$  \\ \cline{2-5}
                      & $5000,20000,2000$   & 26.80     & 100                & 448.263 \\
\hline
\multirow{2}{*}{SPDC} & $1000,5000,500$    & 3.76      & 100                                 & 117.518 \\ \cline{2-5}
                      & $5000,20000,2000$   & 70.10     & 100                & 473.806 \\                      
\hline
\multirow{2}{*}{SP-BCD} & $1000,5000,500$  & $\bold{0.70}$   & $\bold{30}$          & $\bold{111.318}$   \\ \cline{2-5}
                      & $5000,20000,2000$   & $\bold{13.32}$     & $\bold{30}$      & $\bold{448.263}$ \\
\hline
\end{tabular}}
\label{tab:lasso}
\vspace{-2mm}

\end{table}

\subsection{Lasso}
\label{subsec:lasso}
Lasso is an important $l_1$ regularized linear regression, solving the optimization problem,
\begin{align}
\min_{\xB} \frac{1}{2} \| \AB \xB - \bB \|_2^2 + \lambda \| \xB \|_1  \label{eq:lasso}
\end{align}
where $\lambda$ is a regularization parameter, and $\AB \in \Rbb^{m \times n} $ is an observed feature matrix. In typical applications, there are many more features than number of training examples, i.e., $m<n$. By dualizing the first quadratic loss function in (\ref{eq:lasso}), we can have its Sep-CCSP form
\begin{equation}
\min_{\xB \in \Rbb^n} \max_{\yB \in \Rbb^m}\lambda \| \xB \|_1 + \langle \yB, \AB \xB \rangle -  \sum_{i=1}^m \left( \frac{1}{2} y_i^2 + b_i y_i \right). 
\end{equation}
Since $\| \xB \|_1$ is totally separable and non-strongly convex, we can apply our SP-BCD method to the above saddle point problem, i.e., in each iteration we randomly select $K$ coordinates of primal variable $\xB$ to update. For the dual update, the corresponding problem has a simple close-formed solution that can be updated directly.

Due to the vast literature for the Lasso problem, we only choose several representative methods to compare with our method, (1) ISTA (Iterative Shrinkage-Thresholding Algorithm); (2) FISTA (Fast ISTA, \cite{beck2009}); (3) ADMM \cite[Chap 6.4]{boyd2011}, note that the formulation of ADMM for Lasso problem is different from Eq.(\ref{eq:lasso}). ADMM splits the loss function and regularization term using two separable variables, which needs to solve a linear system in each iteration. When the problem size is very large, the time complexity is high and even computationally inacceptable. (4) PDCP~\cite{chambolle2011}, which needs estimation of norm of matrix $\AB$. (5) SPDC~\cite{zhang2015} needs an extra regularization parameter to adapt non-strong convexity. We choose optimal regularization parameter by post-hoc selection. 

We generate the data as in \cite[Chap 11.1]{boyd2011}:
each element of matrix $\AB$, $a_{ij} \sim N(0,1)$ and then normalize the columns to have unit $l_2$ norm; a ``true" value $\xB_{\text{true}} \in \Rbb ^n $ has $d$ nonzeros entries, each of which is sampled from $N(0,1)$; the label $\bB = \AB \xB_{\text{true}} + \epsilon$, where $\epsilon \sim N\left(0,10^{-3}\IB\right)$. The regularization parameter is set as $\lambda = 0.1 \| \AB^T \bB\|_{\infty}$.
The implementation of ISTA, FISTA and ADMM is based on: \url{http://web.stanford.edu/~boyd/papers/prox_algs/lasso.html}. The proximal parameter for these methods are set as $1$.
For SP-BCD and SPDC, we randomly choose $K=100$ coordinates per iteration to run the experiments.

Table \ref{tab:lasso} reports the performance of all these methods on two problems with different sizes and  sparsity. \cut{Figure \ref{fig:lasso} depicts the objective evolution as a function of number of passes and time for problem size: $m=5000, n=20000$ and number of nonzero entries  of $\xB_{\text{true}}$, $d = 500$. }We can observe that SP-BCD uses the least number of passes and time to achieve same objective value with other methods. For smaller sized problems, ADMM also performs very well. However, when the problem size is rising, the computational burden from solving large linear systems becomes a serious issue for ADMM. The issue of scalability also influences the performance of PDCP since it needs the estimation of norm of matrix $\AB$.  Our method SP-BCD is not restricted heavily by a large problem size. SPDC~\cite{zhang2015} even with optimal regularization parameter (by post-hoc selection) still dramatically deteriorates its performance.

\begin{figure*}[ht!]
\vskip -0.1in
\begin{center}
\begin{tabular}{c | ccc}
 \small{$\lambda$} & {\small Objective w.r.t. number of passes} & {\small Objective w.r.t. time} & {\small SP-BCD with different $K$ values}\\
\hline
\small{$10^{-4}$} & \raisebox{-.5\totalheight}{\includegraphics[width=0.26\textwidth]{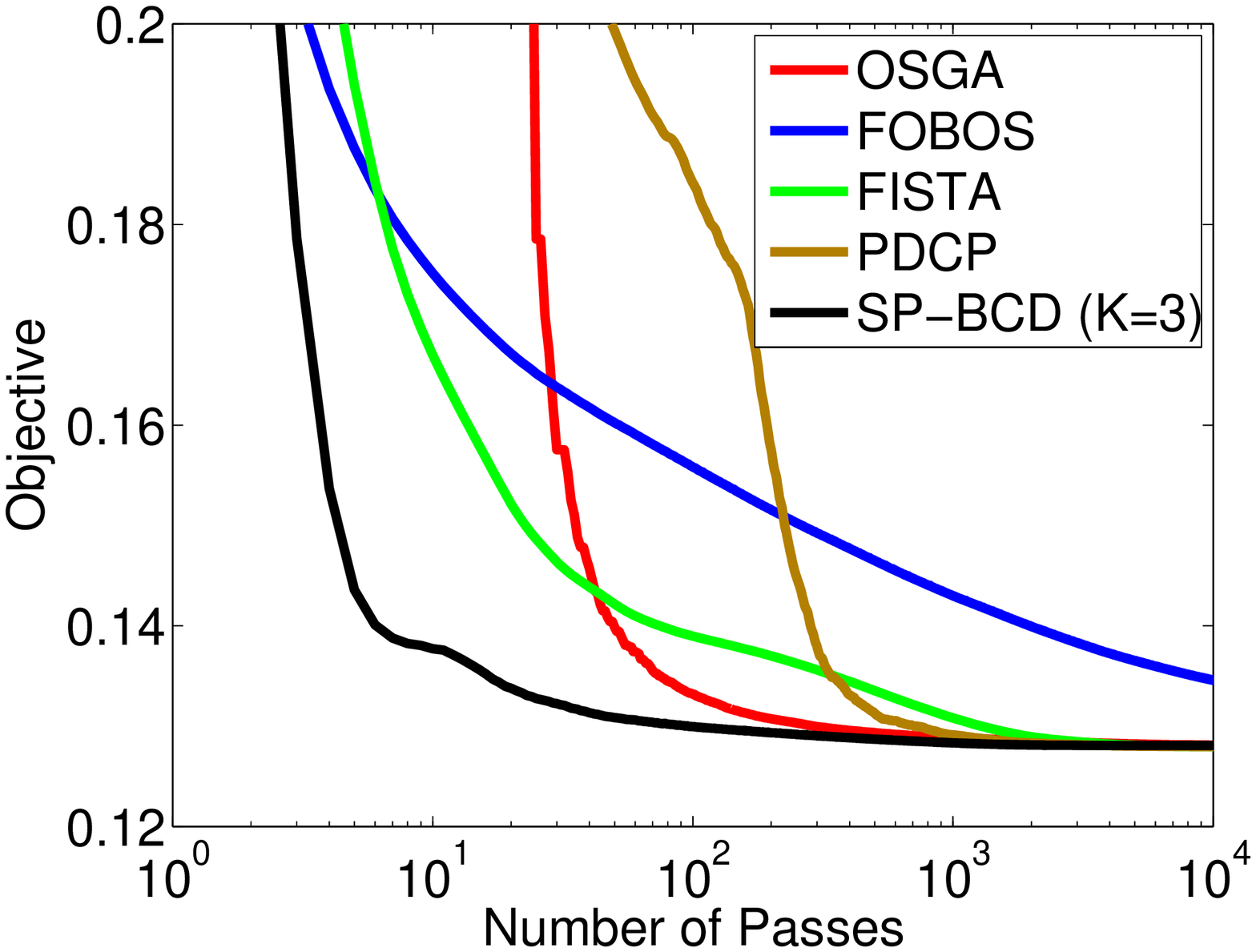}} & \raisebox{-.5\totalheight}{\includegraphics[width=0.26\textwidth]{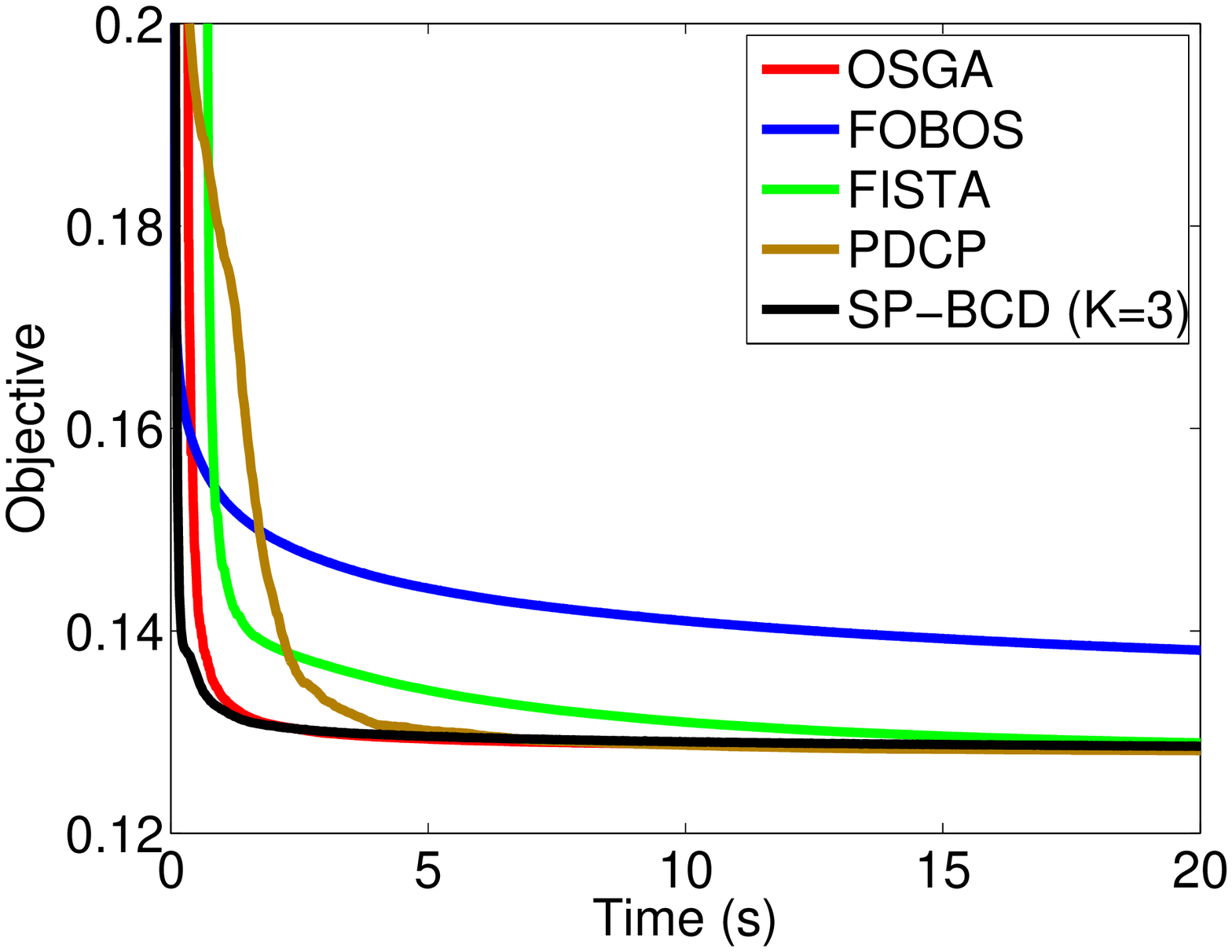}} & \raisebox{-.5\totalheight}{\includegraphics[width=0.26\textwidth]{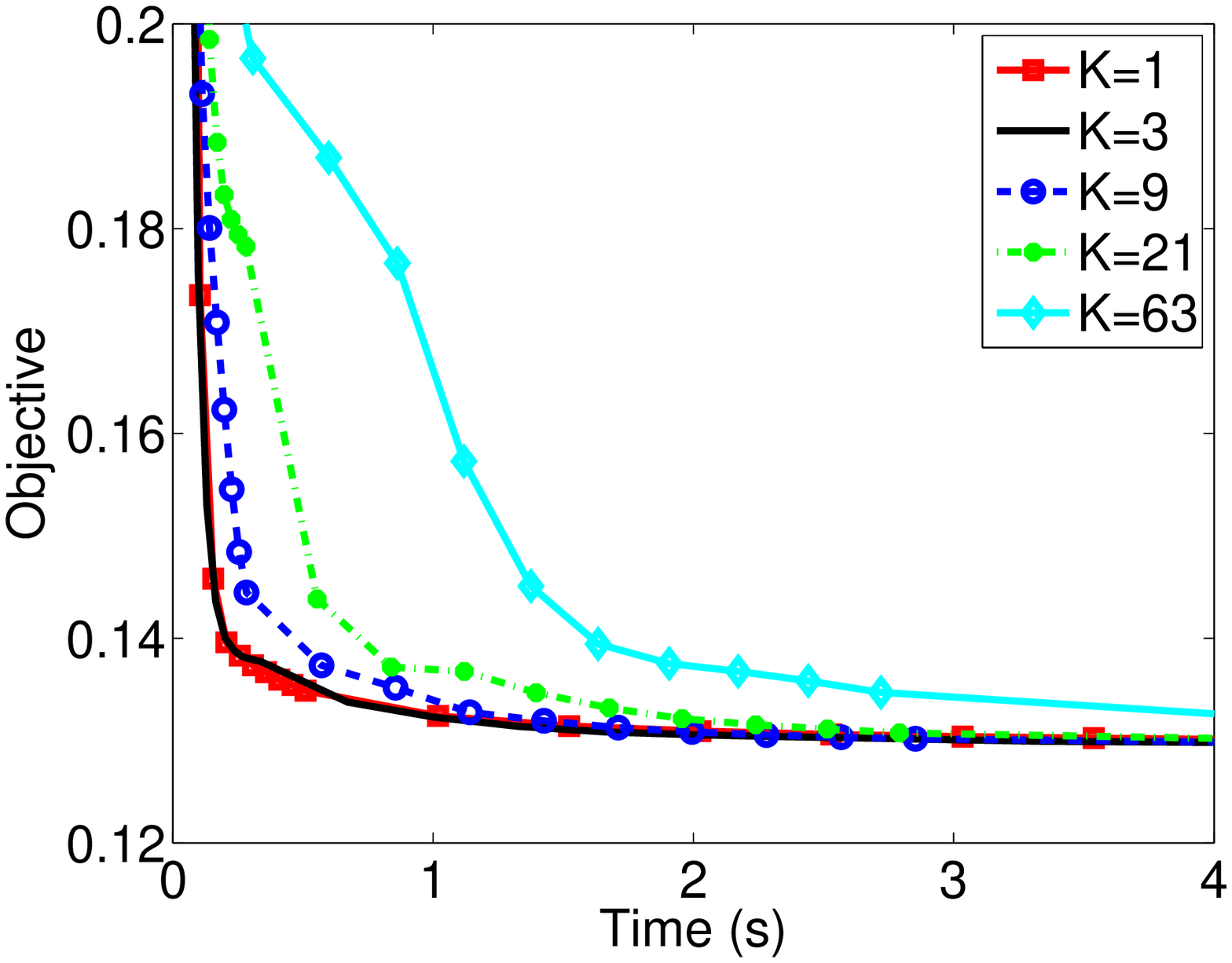}}\\
\small{$10^{-5}$} & \raisebox{-.5\totalheight}{\includegraphics[width=0.26\textwidth]{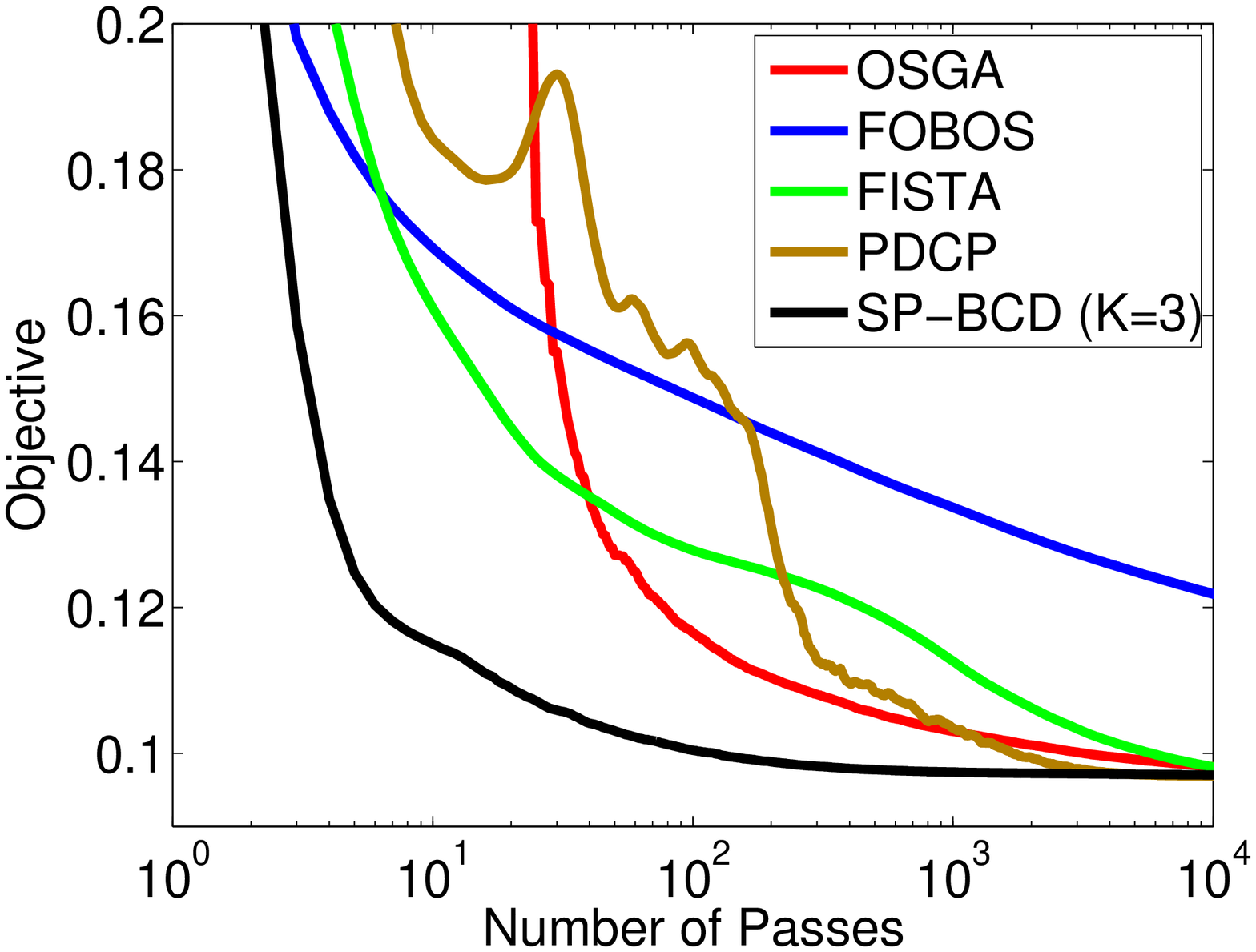}} & \raisebox{-.5\totalheight}{\includegraphics[width=0.26\textwidth]{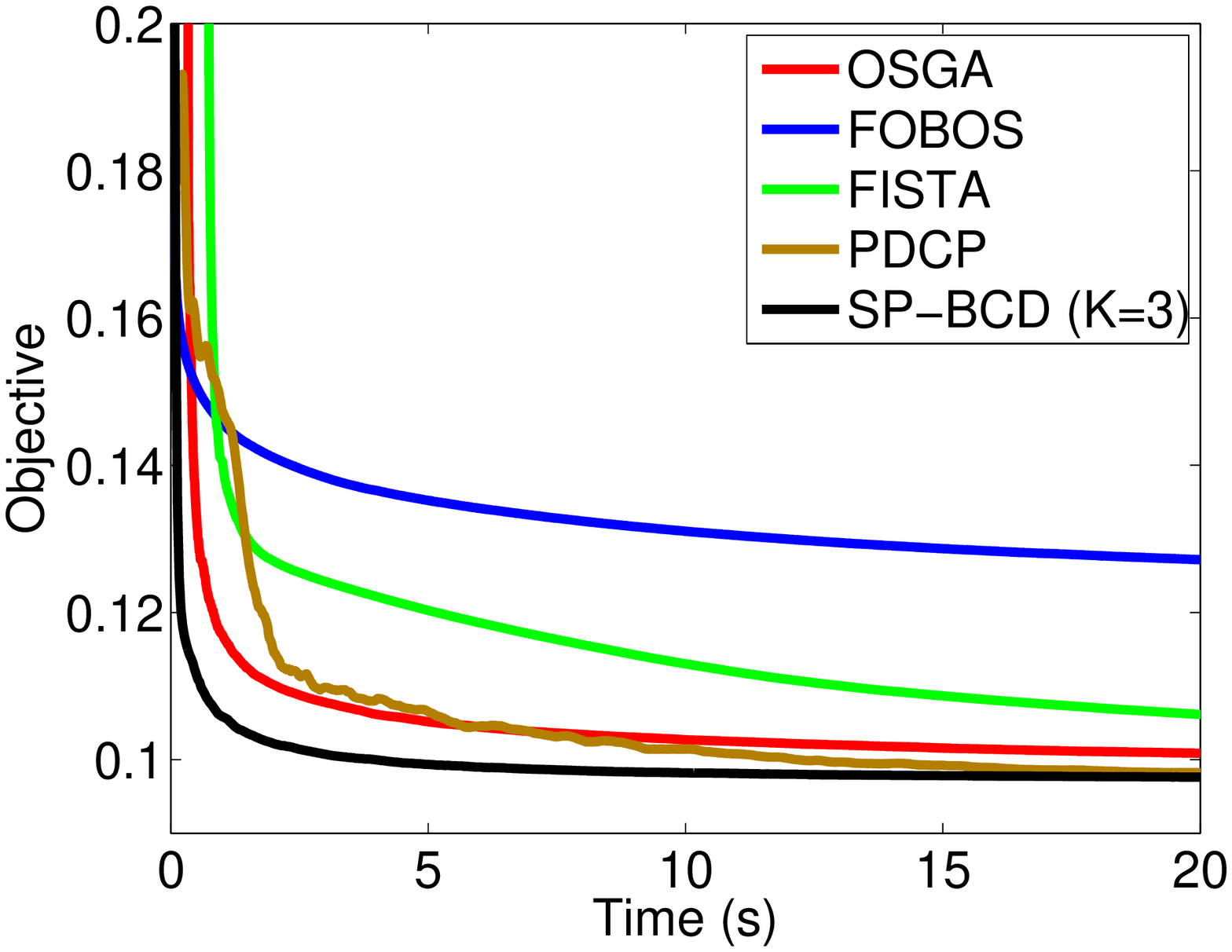}} & \raisebox{-.5\totalheight}{\includegraphics[width=0.26\textwidth]{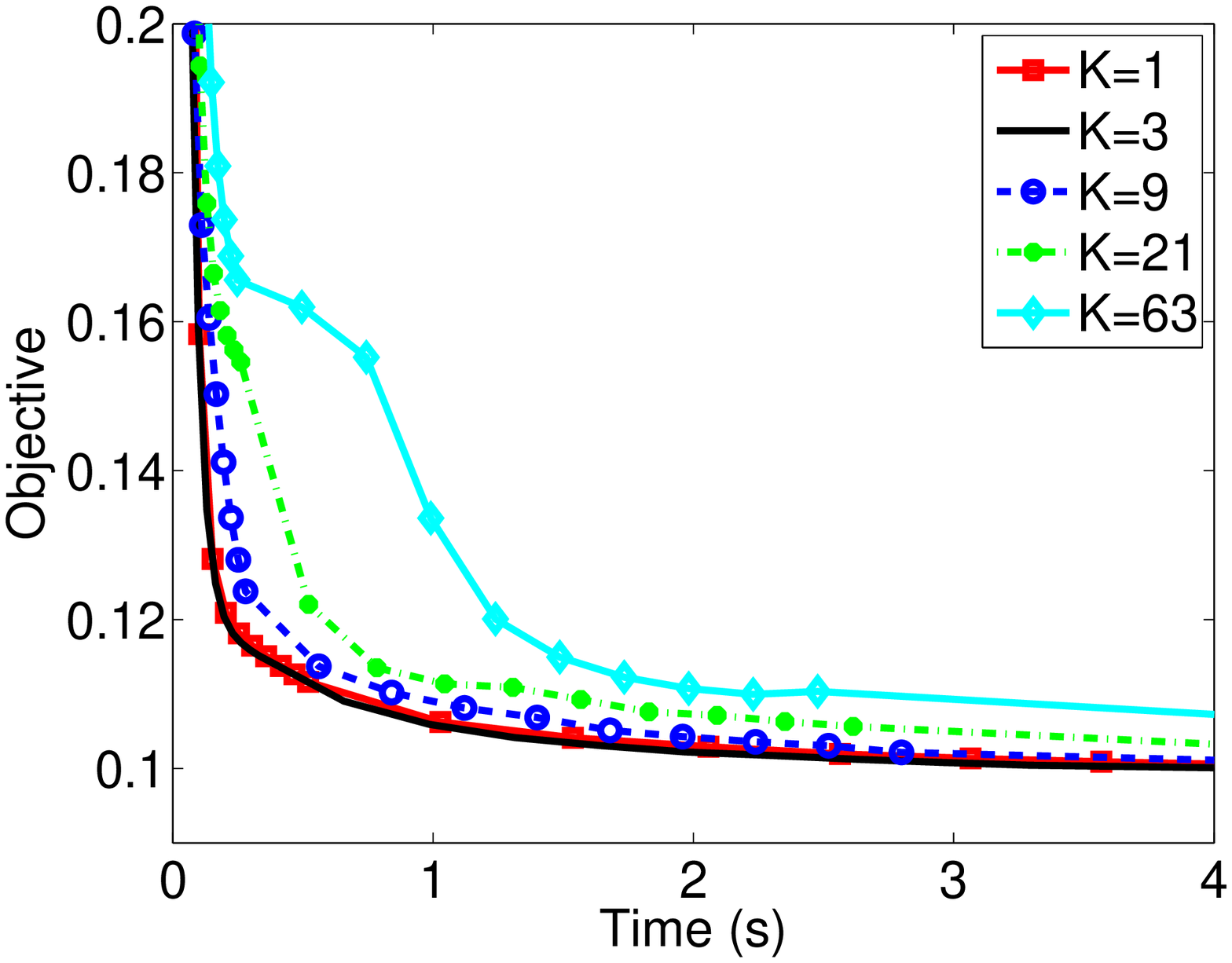}}\\
\small{$10^{-6}$} & \raisebox{-.5\totalheight}{\includegraphics[width=0.26\textwidth]{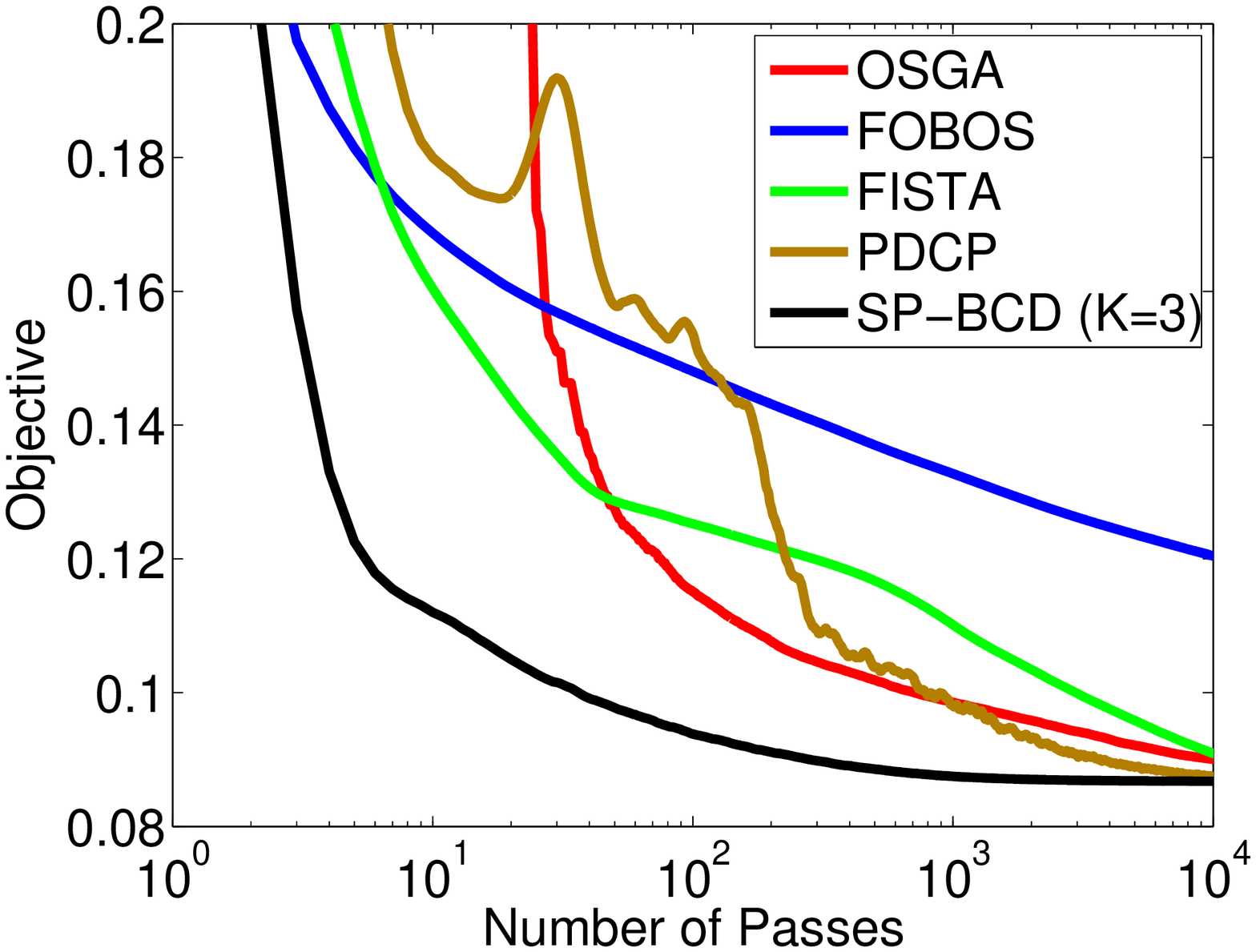}}&\raisebox{-.5\totalheight}{\includegraphics[width=0.26\textwidth]{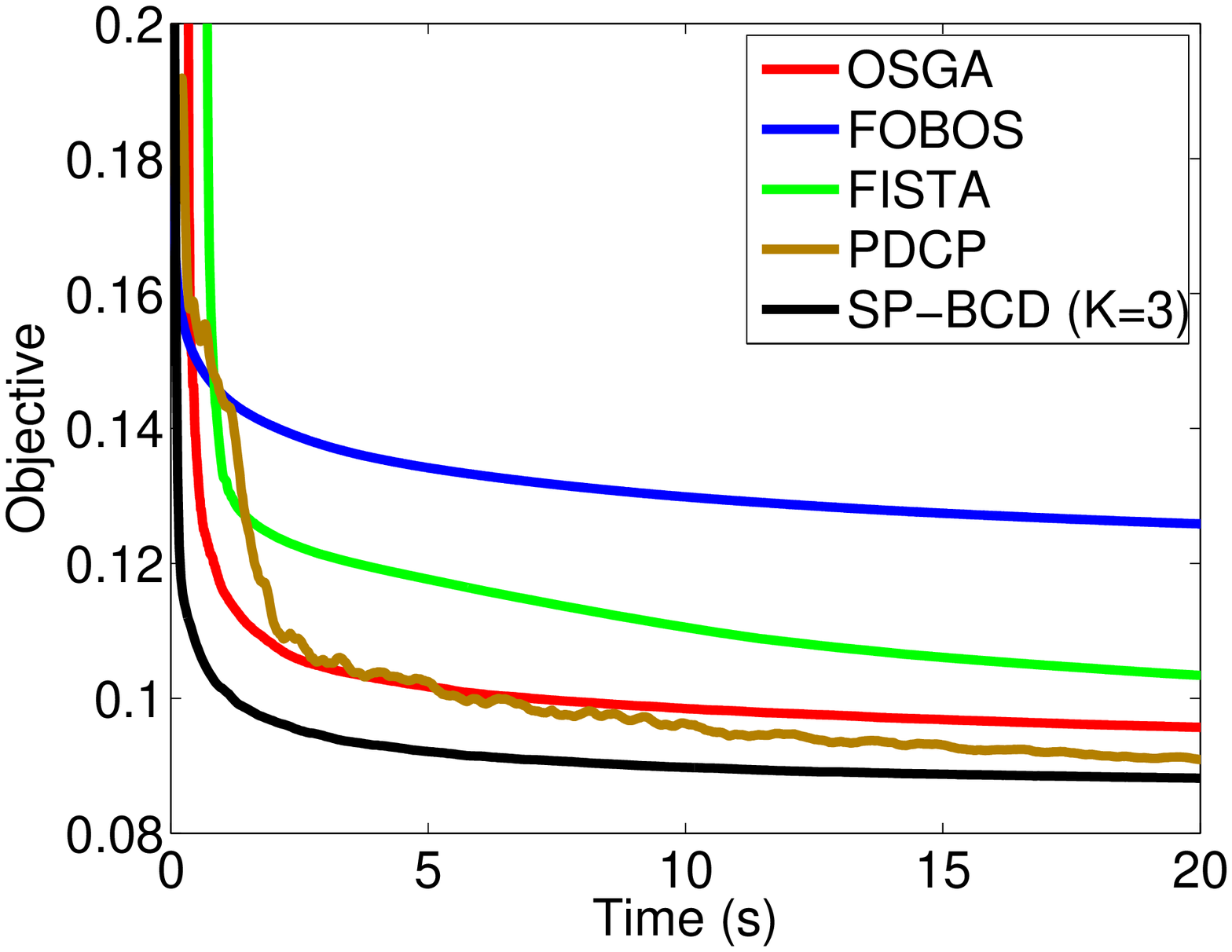}} & \raisebox{-.5\totalheight}{\includegraphics[width=0.26\textwidth]{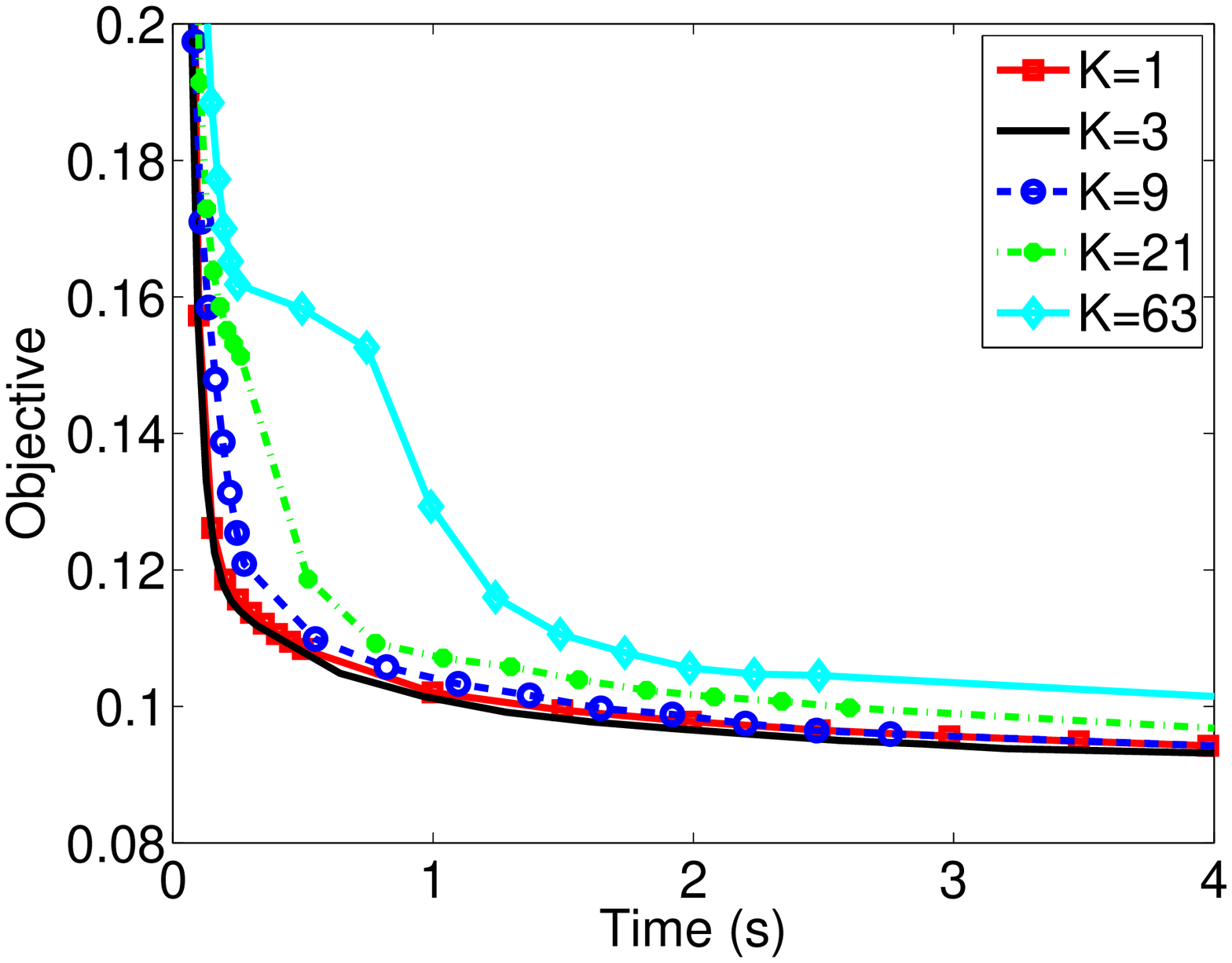}}
\end{tabular}
\vspace{-1mm}
\end{center}
\caption{\small Group Lasso on MEMset dataset with different regularization parameter $\lambda$. 
\label{fig:groupLassoObj}}
\vspace{-2mm}
\end{figure*}

\subsection{Feature Selection with Group Lasso}


We consider solving the following group Lasso problem~\cite{yuan2006}:
\begin{equation}
\min_{\xB} \lambda \sum_{g=1}^G \sqrt{d_g} \| \xB_g \|_2 + \frac{1}{N} \sum_{i=1}^N g_i(\aB_i^T \xB, z_i), \label{eq:groupLasso}
\end{equation}
where $\xB$ is partitioned according to feature grouping, i.e., $\xB = [\xB_1^T, \xB_2^T, \dots, \xB_G^T ]^T$,  each $\aB_i$ is $d$-dimensional feature vector, $z_i \in \{-1, 1 \}$ is the label, and $g_i(\aB_i^T \xB, z_i)$ is a convex loss function, such as the squared loss, logit loss, or hinge loss.  The regularizer is the sum of groupwise $L_2$-norm $\| \xB_g \|_2$, and the trade-off constant $\lambda$ is to balance between the loss and the regularization term. The value $d_g$ accounts for the varying group sizes.
We use hinge loss function $g_i(\aB_i^T \xB, z_i) = \max (0, 1 - z_i\aB_i^T \xB)$ for demonstration. By the conjugate dual transformation of hinge loss,
\begin{equation}
g_i(\aB_i^T \xB, z_i)  = \max_{y_i \in [0,1]} \langle - y_i z_i \aB_i,  \xB \rangle + y_i,
\end{equation}
we can transform the group Lasso problem into the following saddle point problem,
\begin{align}
\min_{\xB} \max_{\yB \in [0,1]^N} \lambda \sum_{g=1}^G \sqrt{d_g} \| \xB_g \|_2 + \frac{1}{N} \langle -\sum_{i=1}^N y_i z_i \aB_i,  \xB \rangle
                                                                                                      + \frac{1}{N}\sum_{i=1}^N y_i
\end{align}
This reformulation of group Lasso makes both the dual and primal update extremely simple and efficient, both of which have closed-formed solution and can be easily derived.

To evaluate the performance of our method for the group Lasso problem, we apply it to a real-world dataset for splice site detection, which plays an important role in gene finding. The MEMset Donor dataset is widely used to demonstrate the advantages of the group Lasso models \cite{meier2008,roth2008}. 
From the original training set, we construct a balanced training set with $8,415$ true and $8,415$ false donor sites. Group lasso on this data with up to 2nd order interactions and up to 4 order interactions
has been analyzed by \cite{meier2008,roth2008}, respectively. As shown in \cite{roth2008}, there is not much improvement using higher order
interactions. Therefore we only consider all
three-way and lower order interactions. This forms $G=63$
groups or $d = 2604$-dimensional feature space with $\{ 7, 21, 35 \}$ groups of $\{4, 16, 64\}$-dimensional coordinate block, respectively.

We compare our SP-BCD with several recent competitive optimization methods for the non-smooth regularized problem: (1) OSGA~\cite{neumaier2014osga}, a fast subgradient algorithm
with optimal complexity; (2) FOBOS~\cite{duchi2009} based on Forward-Backward splitting; (3) FISTA~\cite{beck2009}, using a smoothing technique to make it applicable with smoothing parameter $\epsilon = 5 \times 10^{-4}$; (4) PDCP~\cite{chambolle2011}.

In this application, we evaluate the performance of these methods under different regularization parameter $\lambda = \{ 10^{-4}, 10^{-5}, 10^{-6} \}$. The first two columns in Figure \ref{fig:groupLassoObj} compares our method SP-BCD (with $K=3$) with other methods in terms of the evolution of the objective function in Eq.(\ref{eq:groupLasso}) both w.r.t. the number of passes and w.r.t time. In all these test cases, SP-BCD demonstrates its superiority on both number of passes and consumed time. When the regularization is strong with large $\lambda = 10^{-4}$, all the methods tend to converge fast, but SP-BCD is the fastest one. PDCP performs poorly in first hundreds or thousands of passes, since it only applies the constant stepsize $1/\| A \|$. Compared with PDCP, our method considers the structure of matrix $\AB$ and scales each dimension of primal and dual updates, which can achieve better empirical performance.

In order to investigate the effect of the number of chosen blocks for our method, we implement it using different $K$ values, $K = \{ 1, 3, 9,21, 63 \}$. The results are shown in the third column of Figure \ref{fig:groupLassoObj}. In all the tested cases, a smaller number of blocks yields faster convergence, which shows the advantage of the flexible stochastic update of our method compared with~\cite{pock2011}.

\section{Conclusion and Future Work}
\label{sec:con}
We propose a Stochastic Parallel Block Coordinate Descent (SP-BCD) for the Sep-CCSP problem, especially for non-strongly convex functions. SP-BCD shares the efficiency and flexibility of block coordinate descent methods while keeping the simplicity of primal-dual methods and utilizing the structure of matrix $\AB$. Many machine learning applications are covered and we compare SP-BCD with other competitive methods in each application, showing the benefits of SP-BCD over others on robust PCA, Lasso and group Lasso.  An immediate future direction is to investigate other valid parameter configurations.

\section*{Appendix: Proof of Theorem 1}
Firstly, we analyze the primal and dual variables $\xB$ and $\yB$ after $t$-th update in the Algorithm \ref{alg:SP-BCD}. We introduce a  temporary variable $\tilde{\xB}_j$ to be the value of $\xB_j^{t+1}$ if $j \in S_t$, for any $j \in \{1, 2, \dots, J\}$, i.e. (crossref Eq.(\ref{eq:primalupdate})),
\begin{equation}
\tilde{\xB}_j = \argmin_{\xB_j} f_j(\xB_j) + \langle \yB^t, \AB_j \xB_j \rangle + \frac{1}{2} \| \xB_j - \xB_j^t \|_{\hB_j}^2
\end{equation}
Due to the strong convexity of the added  proximal term, the function minimized above is $\hB_j$-strongly convex, and then for any $\xB_j$ we have
\begin{equation}
f_j(\xB_j) + \langle \yB^t, \AB_j \xB_j\rangle + \frac{1}{2} \| \xB_j - \xB_j^t \|_{\hB_j}^2 \geq
f_j(\tilde{\xB}_j) + \langle \yB^t, \AB_j \tilde{\xB}_j\rangle + \frac{1}{2} \| \tilde{\xB}_j - \xB_j^t \|_{\hB_j}^2 
+ \frac{1}{2} \| \tilde{\xB}_j - \xB_j \|_{\hB_j}^2.  \label{eq:primal_jnequality1}
\end{equation}
In our algorithm, an index set $S_t$ is randomly chosen. For every specific index $j$, the event $j \in S_t$ happens with probability $K/J$. If $j \in S_t$, then $\xB_j^{t+1}$ is updated to the value $\tilde{\xB}_j^t$. Otherwise, $\xB_j^{t+1}$ is kept to be its old value $\xB_j^t$. Let $\xi_t$ be the random event that contains the set of all random variable before round $t$,
\begin{equation}
\xi_t = \{S_1, S_2, \dots, S_t  \},
\end{equation}
and then we have
\begin{align*}
\Ebb_{\xi_t} \left[ \| \xB_j^{t+1} - \xB_j \|_{\hB_j}^2  \right] &= \frac{K}{J} \| \tilde{\xB}_j - \xB_j \|_{\hB_j}^2 + \frac{J-K}{J} \| \xB_j^{t} - \xB_j \|_{\hB_j}^2  \\
\Ebb_{\xi_t} \left[ \| \xB_j^{t+1} - \xB_j^t\|_{\hB_j}^2  \right] &= \frac{K}{J} \| \tilde{\xB}_j - \xB_j^t \|_{\hB_j}^2\\
\Ebb_{\xi_t} \left[ \xB_j^{t+1}  \right] &= \frac{K}{J} \tilde{\xB}_j + \frac{J-K}{J} \xB_j^t  \\
\Ebb_{\xi_t} \left[ f_j ( \xB_j^{t+1} )  \right] &= \frac{K}{J} f_j ( \tilde{\xB}_j)  + \frac{J-K}{J} f_j(\xB_j^t)
\end{align*}
With these equality relationships, we can substitute $f_j\left(\tilde{\xB}_j \right)$, $\tilde{\xB}_j$, $\| \tilde{\xB}_j - \xB_j^t \|_{\hB_j}^2$ and $\| \tilde{\xB}_j - \xB_j \|_{\hB_j}^2$ into the inequality (\ref{eq:primal_jnequality1}),
\begin{align*}
\Ebb_{\xi_t} \left[  f_j(\xB_j^{t+1})  \right] - f_j(\xB_j) &\leq  \left( \frac{J}{K}\cdot \frac{1}{2} \| \xB_j^t - \xB_j \|_{\hB_j}^2 + \frac{J-K}{K} f_j(\xB_j^t)\right) - \left( \frac{J}{K} \cdot \frac{1}{2}\Ebb_{\xi_t} \left[  \| \xB_j^{t+1} - \xB_j  \|_{\hB_j}^2  \right] +  \frac{J-K}{K} f_j(\xB_j^{t+1}) \right) \\
& - \frac{J}{K} \cdot \frac{1}{2} \Ebb_{\xi_t} \left[  \| \xB_j^{t+1} - \xB_j^t  \|_{\hB_j}^2  \right] - \Ebb_{\xi_t} \left[  \langle \yB^t, \AB_j \left(  \frac{J}{K}\xB_j^{t+1} - \frac{J-K}{K} \xB_j^t - \frac{K}{J}\xB_j  \right) \rangle  \right].
\end{align*}

Then summing the above inequality with all the indices $i = 1, \dots, J$, we can obtain
\begin{align}
\Ebb_{\xi_t} \left[  f(\xB^{t+1})  \right] - f(\xB) &\leq  \left( \frac{J}{K}\cdot \frac{1}{2} \| \xB^t - \xB \|_{\hB}^2 + \frac{J-K}{K} f(\xB^t)\right)  - \left( \frac{J}{K} \cdot \frac{1}{2}\Ebb_{\xi_t} \left[  \| \xB^{t+1} - \xB  \|_{\hB}^2  \right] +  \frac{J-K}{K} f(\xB^{t+1}) \right) \nonumber \\
& - \frac{J}{K} \cdot \frac{1}{2} \Ebb_{\xi_t} \left[  \| \xB^{t+1} - \xB^t  \|_{\hB}^2  \right]  - \Ebb_{\xi_t} \left[  \langle \yB^t, \AB \left(  \frac{J}{K}\xB^{t+1} - \frac{J-K}{K} \xB^t - \frac{K}{J}\xB  \right) \rangle  \right]. \label{eq:primal_jnequality2}
\end{align}

Now, we  consider dual update in Eq. (\ref{eq:dualupdate}),
\begin{align}
g^*(\yB) - \langle \yB, \overline{\rB}^t + \frac{J}{K} \sum_{j \in S_t} \AB_j \left( \overline{\xB}_j^{t+1} - \overline{\xB}_j^t  \right)  \rangle + \frac{1}{2} \|  \yB - \yB^t  \|_{\sigmaB^t}^2 & \geq g^*(\yB^{t+1}) - \langle \yB^{t+1}, \overline{\rB}^t + \frac{J}{K} \sum_{j \in S_t} \AB_j \left( \overline{\xB}_j^{t+1} - \overline{\xB}_j^t  \right)  \rangle \nonumber \\
& \quad + \frac{1}{2} \|  \yB^{t+1} - \yB^t  \|_{\sigmaB^t}^2 + \frac{1}{2} \| \yB^{t+1} - \yB \|_{\sigmaB^t}^2 \label{eq:dual_jnequality1}
\end{align}
Since in each iteration, we always keep $\overline{\rB}^t  = \sum_{i=1}^J \AB_j \overline{\xB}_j^{t}$, thus we have
\begin{align}
\Ebb_{\xi_t} \left[ \overline{\rB}^t + \frac{J}{K} \sum_{j \in S_t} \AB_j \left( \overline{\xB}_j^{t+1} - \overline{\xB}_j^t  \right) \right] &=  \Ebb_{\xi_t} \left[ \overline{\rB}^t + \frac{J}{K} \sum_{i =1}^J \AB_j \left( \overline{\xB}_j^{t+1} - \overline{\xB}_j^t  \right) \right] \nonumber \\
&=\Ebb_{\xi_t} \left[ \AB \left( \frac{J}{K}\overline{\xB}^{t+1} - \frac{J-K}{K}\overline{\xB}^{t}  \right) \right] \label{eq:crossmorph}
\end{align}
Considering the intermediate variable $\overline{\xB}_j^{t+1}$ in Eq.(\ref{eq:extrapolation}), we have
\begin{align*}
\Ebb_{\xi_t} \left[  \overline{\xB}_j^{t+1}  \right] &= \frac{K}{J} \left(  \tilde{\xB}_j^{t+1} + \theta \left( \tilde{\xB}_j^{t+1} - \xB_j^t \right)  \right) + \frac{J-K}{J} \overline{\xB}_j^t \nonumber \\
&= \frac{K}{J} \left(  \frac{J}{K} \Ebb_{\xi_t} \left[ \xB_j^{t+1} \right] - \frac{J-K}{K}\xB_j^t + \theta \left( \frac{J}{K} \Ebb_{\xi_t} \left[ \xB_j^{t+1} \right] - \frac{J-K}{K}\xB_j^t - \xB_j^t \right)  \right) + \frac{J-K}{J} \overline{\xB}_j^t \nonumber \\
&= \Ebb_{\xi_t} \left[ \xB_j^{t+1} \right] + \theta \left(  \Ebb_{\xi_t} \left[ \xB_j^{t+1} \right] - \xB_j^t \right) + \frac{J-K}{J} \left(  \overline{\xB}_j^t - \xB_j^t  \right)
\end{align*}
Given the parameter $\theta = \frac{K}{J}$, then
\begin{equation*}
\Ebb_{\xi_t} \left[  \overline{\xB}^{t+1}  \right] = \left(1+ \frac{K}{J}\right) \Ebb_{\xi_t} \left[  \xB^{t+1}  \right] + \left( 1 - \frac{2K}{J} \right)\xB^t + \left( 1 - \frac{K}{J} \right) \overline{\xB}^t
\end{equation*}
Plugging the above equality into the inequality (\ref{eq:crossmorph}),
\begin{align}
\Ebb_{\xi_t} \left[ \overline{\rB}^t + \frac{J}{K} \sum_{j \in S_t} \AB_j \left( \overline{\xB}_j^{t+1} - \overline{\xB}_j^t  \right) \right] = \Ebb_{\xi_t} \left[ \AB \left( \frac{J+K}{K} \xB^{t+1} - \frac{J}{K}\xB^t  \right) \right]
\end{align}
We assign expectation to both sides of the inequality (\ref{eq:dual_jnequality1}) and plug in $\Ebb_{\xi_t} \left[ \overline{\rB}^t + \frac{J}{K} \sum_{j \in S_t} \AB_j \left( \overline{\xB}_j^{t+1} - \overline{\xB}_j^t  \right) \right]$, and after some manipulations,
\begin{align}
\Ebb_{\xi_t} \left[ g^*(\yB^{t+1}) \right] - g^*(\yB)  \leq & \frac{1}{2} \|  \yB^t - \yB  \|_{\sigmaB}^2 - \frac{1}{2} \Ebb_{\xi_t} \left[ \| \yB^{t+1} - \yB \|_{\sigmaB^t}^2 \right] - \frac{1}{2} \Ebb_{\xi_t} \left[ \| \yB^{t+1} - \yB^t \|_{\sigmaB^t}^2 \right]  \nonumber \\
&+ \Ebb_{\xi_t} \left[ \langle \yB^t - \yB, \AB \left( \frac{J+K}{K} \xB^{t+1} - \frac{J}{K}\xB^t  \right) \rangle \right]. \label{eq:dual_jnequality2}
\end{align}
Now we are ready to use the two key inequalities (\ref{eq:primal_jnequality2}) and (\ref{eq:dual_jnequality2}) to construct the following gap
\begin{align}
\Ebb_{\xi_t} \left[ L(\xB^{t+1}, \yB) - L(\xB, \yB^{t+1})  \right] & = \Ebb_{\xi_t} \left[ f(\xB^{t+1}) + \langle \yB, \AB \xB^{t+1} \rangle \right] - g^*(\yB)  - \left( f(\xB) +  \Ebb_{\xi_t} \left[ \langle \yB^{t+1}, \AB \xB \rangle - g^*(\yB^{t+1}) \right] \right) \nonumber \\
&= \Ebb_{\xi_t} \left[  f(\xB^{t+1})  \right] - f(\xB) + \Ebb_{\xi_t} \left[ g^*(\yB^{t+1}) \right] - g^*(\yB)
+ \Ebb_{\xi_t} \left[ \langle \yB, \AB \xB^{t+1} \rangle  - \langle \yB^{t+1}, \AB \xB \rangle \right] \\
&\leq
 \left( \frac{J}{K}\cdot \frac{1}{2} \| \xB^t - \xB \|_{\hB}^2 + \frac{1}{2} \|  \yB^t - \yB  \|_{\sigmaB^t}^2 + \frac{J-K}{K} f(\xB^t) \right)  \nonumber \\
&- \left( \frac{J}{K} \cdot \frac{1}{2}\Ebb_{\xi_t} \left[  \| \xB^{t+1} - \xB  \|_{\hB}^2  \right] + \frac{1}{2} \Ebb_{\xi_t} \left[ \| \yB^{t+1} - \yB \|_{\sigmaB^t}^2 \right] + \frac{J-K}{K} f(\xB^{t+1}) \right) \nonumber \\
&- \left( \frac{J}{K} \cdot \frac{1}{2} \Ebb_{\xi_t} \left[  \| \xB^{t+1} - \xB^t  \|_{\hB}^2  \right] + \frac{1}{2} \Ebb_{\xi_t} \left[ \| \yB^{t+1} - \yB^t \|_{\sigmaB^t}^2 \right] \right)  \nonumber \\
&- \Ebb_{\xi_t} \left[  \langle \yB^t, \AB \left(  \frac{J}{K}\xB^{t+1} - \frac{J-K}{K} \xB^t - \frac{K}{J}\xB  \right) \rangle  \right] \nonumber \\
&+ \Ebb_{\xi_t} \left[ \langle \yB^t - \yB, \AB \left( \frac{J+K}{K} \xB^{t+1} - \frac{J}{K}\xB^t  \right) \rangle \right] + \Ebb_{\xi_t} \left[ \langle \yB, \AB \xB^{t+1} \rangle  - \langle \yB^{t+1}, \AB \xB \rangle \right].
\end{align}
After some sophisticated manipulations and rearrangements of the R.H.S of the above inequality, we can obtain
\begin{equation}
\Ebb_{\xi_t} \left[ L(\xB^{t+1}, \yB) - L(\xB, \yB^{t+1})  \right] \leq M(t) - M(t+1) - C(t,t+1), \label{eq:key1}
\end{equation}
where
\begin{multline}
M(t) = \frac{J}{K}\cdot \frac{1}{2} \| \xB^t - \xB \|_{\hB}^2 + \frac{1}{2} \|  \yB^t - \yB  \|_{\sigmaB^t}^2 - \langle  \yB^t - \yB, \AB \left( \xB^t - \xB \right) \rangle 
+\frac{J-K}{K} \left( f(\xB^t) + \langle \yB, \AB \xB^t \rangle - \left( f(\xB) +   \langle \yB, \AB \xB  \rangle  \right)    \right),
\end{multline}
and
\begin{align}
M(t+1)&= \frac{J}{K} \cdot \frac{1}{2}\Ebb_{\xi_t} \left[  \| \xB^{t+1} - \xB  \|_{\hB}^2  \right] + \frac{1}{2} \Ebb_{\xi_t} \left[ \| \yB^{t+1} - \yB \|_{\sigmaB^t}^2 \right]- \Ebb_{\xi_t} \left[ \langle  \yB^{t+1} - \yB, \AB \left( \xB^{t+1} - \xB \right) \rangle \right]  \nonumber \\
&+\frac{J-K}{K} \Ebb_{\xi_t} \left[ \left( f(\xB^{t+1}) + \langle \yB, \AB \xB^{t+1} \rangle - \left( f(\xB) +   \langle \yB, \AB \xB  \rangle  \right)    \right) \right],
\end{align}
and
\begin{align}
C(t, t+1) &= \frac{J}{K} \cdot \frac{1}{2} \Ebb_{\xi_t} \left[ \| \xB^{t+1} - \xB^{t} \|_{\hB}^2  \right] + \frac{1}{2} \Ebb_{\xi_t} \left[ \| \yB^{t+1} - \yB^{t} \|_{\sigmaB^t}^2  \right] 
- \frac{J}{K} \Ebb_{\xi_t} \left[ \langle  \yB^{t+1} - \yB^t, \AB \left( \xB^{t+1} - \xB^t \right) \rangle \right] \\
&= \frac{J}{K} \cdot \frac{1}{2} \Ebb_{\xi_t} \left[ \sum_{j \in S_t} \|  \xB_j^{t+1} - \xB_j^{t} \|_{\hB}^2  \right] + \frac{1}{2} \Ebb_{\xi_t} \left[ \| \yB^{t+1} - \yB^{t} \|_{\sigmaB^t}^2  \right] - \frac{J}{K} \Ebb_{\xi_t} \left[ \langle  \yB^{t+1} - \yB^t, \sum_{j \in S_t}\AB_j \left( \xB_j^{t+1} - \xB_j^t \right) \rangle \right].
\end{align}
Now we bound the term $C(t, t+1)$ given the following parameter configuration,
\begin{align*}
\quad h_{d} &= \sum_{j=1}^m |  A_{jd} |, \quad d = 1, 2, \dots, n \\
\sigma_k^t &= \frac{J}{K} \sum_{j \in S_t} | A_{kj} |, \quad k = 1,2,\dots,m.
\end{align*}
We can easily observe that the above parameter configuration makes the following symmetric matrix $\PB$ diagonally dominant, which guarantees its positive semidefiniteness:
\begin{equation}
\PB =
\begin{bmatrix}
\diag (\hB_{S_t}) & -\AB_{S_t}^T \\
-\AB_{S_t} & \frac{K}{J} \diag (\sigmaB^t)
\end{bmatrix}
\succeq 0.
\end{equation}
Therefore, this directly leads $C(t, t+1) \geq 0$, and we can further simplify the inequality (\ref{eq:key1}),
\begin{equation}
\Ebb_{\xi_t} \left[ L(\xB^{t+1}, \yB) - L(\xB, \yB^{t+1})  \right] \leq M(t) - M(t+1),  .
\end{equation}
Summing the above inequality from $t= 0,1,\dots,T-1$ and we  find
\begin{equation}
\Ebb \left[ \sum_{t=1}^T L(\xB^{t}, \yB) - L(\xB, \yB^{t})  \right] \leq M(0) - M(T) \label{eq:key2} 
\end{equation}
Since $(\xB, \yB)$ is a saddle point, for any $t$ we have
\begin{equation}
f(\xB^{t+1}) + \langle \yB, \AB \xB^{t+1} \rangle - \left( f(\xB) +   \langle \yB, \AB \xB  \rangle  \right) \geq 0 \label{eq:ineq_saddle_induced}
\end{equation}

Thanks again to the positive semidefiniteness of the matrix $\PB$ and the inequality (\ref{eq:ineq_saddle_induced}) when $t=T-1$, we have
\begin{equation*}
M(T) \geq 0,
\end{equation*}
which further simplifies inequality (\ref{eq:key2})
\begin{align*}
&\Ebb \left[ \sum_{t=1}^T L(\xB^{t}, \yB) - L(\xB, \yB^{t})  \right]  \leq M(0)
\end{align*}

Finally, applying the convexity of the function $(\xB ', \yB ') \mapsto L(\xB ', \yB) - L(\xB, \yB ') $, we have
\begin{equation}
\Ebb \left[ L\left(\sum_{t=1}^T \xB^t/T, \yB \right) - L\left(\xB, \sum_{t=1}^T \yB^t/T \right)  \right]  \leq  \frac{1}{T} M(0),
\end{equation}


which completes the proof.

\section*{Acknowledgements}

ZZ gratefully acknowledges the financial support from the University of Edinburgh and China Scholarship Council.

\bibliographystyle{plain}
\bibliography{saddle}

\begin{thebibliography}{10}

\bibitem{beck2009}
A.~Beck and M.~Teboulle.
\newblock A fast iterative shrinkage-thresholding algorithm for linear inverse
  problems.
\newblock {\em SIAM Journal on Imaging Sciences}, 2(1):183--202, 2009.

\bibitem{boyd2011}
S.~Boyd, N.~Parikh, E.~Chu, B.~Peleato, and J.~Eckstein.
\newblock Distributed optimization and statistical learning via the alternating
  direction method of multipliers.
\newblock {\em Foundations and Trends{\textregistered} in Machine Learning},
  3(1):1--122, 2011.

\bibitem{candes2011}
E.~J. Cand{\`e}s, X.~Li, Y.~Ma, and J.~Wright.
\newblock Robust principal component analysis?
\newblock {\em Journal of the ACM (JACM)}, 58(3):11, 2011.

\bibitem{chambolle2011}
A.~Chambolle and T.~Pock.
\newblock A first-order primal-dual algorithm for convex problems with
  applications to imaging.
\newblock {\em Journal of Mathematical Imaging and Vision}, 40(1):120--145,
  2011.

\bibitem{chambolle2014}
A.~Chambolle and T.~Pock.
\newblock On the ergodic convergence rates of a first-order primal-dual
  algorithm.
\newblock {\em Optimization-online preprint}, 2014.

\bibitem{chen2001}
S.~Chen, D.~Donoho, and M.~A. Saunders.
\newblock Atomic decomposition by basis pursuit.
\newblock {\em SIAM review}, 43(1):129--159, 2001.

\bibitem{defazio2014saga}
A.~Defazio, F.~Bach, and S.~Lacoste-Julien.
\newblock Saga: A fast incremental gradient method with support for
  non-strongly convex composite objectives.
\newblock In {\em Advances in Neural Information Processing Systems}, pages
  1646--1654, 2014.

\bibitem{duchi2009}
J.~Duchi and Y.~Singer.
\newblock Efficient online and batch learning using forward backward splitting.
\newblock {\em The Journal of Machine Learning Research}, 10:2899--2934, 2009.

\bibitem{hastie2009}
T.~Hastie, R.~Tibshirani, and J.~Friedman.
\newblock {\em The elements of statistical learning}, volume~2.
\newblock Springer, 2009.

\bibitem{hong2012}
M.~Hong and Z.~Luo.
\newblock On the linear convergence of the alternating direction method of
  multipliers.
\newblock {\em arXiv preprint arXiv:1208.3922}, 2012.

\bibitem{meier2008}
L.~Meier, S.~Van De~Geer, and P.~B{\"u}hlmann.
\newblock The group lasso for logistic regression.
\newblock {\em Journal of the Royal Statistical Society: Series B (Statistical
  Methodology)}, 70(1):53--71, 2008.

\bibitem{nesterov2012efficiency}
Y.~Nesterov.
\newblock Efficiency of coordinate descent methods on huge-scale optimization
  problems.
\newblock {\em SIAM Journal on Optimization}, 22(2):341--362, 2012.

\bibitem{neumaier2014osga}
A.~Neumaier.
\newblock Osga: A fast subgradient algorithm with optimal complexity.
\newblock {\em arXiv preprint arXiv:1402.1125}, 2014.

\bibitem{parikh2013}
N.~Parikh and S.~Boyd.
\newblock Proximal algorithms.
\newblock {\em Foundations and Trends in Optimization}, 1(3):123--231, 2013.

\bibitem{parikh2014}
N.~Parikh and S.~Boyd.
\newblock Block splitting for distributed optimization.
\newblock {\em Mathematical Programming Computation}, 6(1):77--102, 2014.

\bibitem{pock2011}
T.~Pock and A.~Chambolle.
\newblock Diagonal preconditioning for first order primal-dual algorithms in
  convex optimization.
\newblock In {\em 2011 IEEE International Conference on Computer Vision
  (ICCV)}, pages 1762--1769. IEEE, 2011.

\bibitem{richtarik2014iteration}
P.~Richt{\'a}rik and M.~Tak{\'a}{\v{c}}.
\newblock Iteration complexity of randomized block-coordinate descent methods
  for minimizing a composite function.
\newblock {\em Mathematical Programming}, 144(1-2):1--38, 2014.

\bibitem{richtarik2015parallel}
P.~Richt{\'a}rik and M.~Tak{\'a}{\v{c}}.
\newblock Parallel coordinate descent methods for big data optimization.
\newblock {\em Mathematical Programming}, pages 1--52, 2015.

\bibitem{roth2008}
V.~Roth and B.~Fischer.
\newblock The group-lasso for generalized linear models: uniqueness of
  solutions and efficient algorithms.
\newblock In {\em Proceedings of the 25th international conference on Machine
  learning}, pages 848--855. ACM, 2008.

\bibitem{wainwright2014}
M.~J. Wainwright.
\newblock Structured regularizers for high-dimensional problems: Statistical
  and computational issues.
\newblock {\em Annual Review of Statistics and Its Application}, 1:233--253,
  2014.

\bibitem{wang2014}
H.~Wang, A.~Banerjee, and Z.~Luo.
\newblock Parallel direction method of multipliers.
\newblock In {\em Advances in Neural Information Processing Systems 27}, pages
  181--189, 2014.

\bibitem{wright2009}
J.~Wright, A.~Ganesh, S.~Rao, Y.~Peng, and Y.~Ma.
\newblock Robust principal component analysis: Exact recovery of corrupted
  low-rank matrices via convex optimization.
\newblock In {\em Advances in neural information processing systems}, pages
  2080--2088, 2009.

\bibitem{yuan2006}
M.~Yuan and Y.~Lin.
\newblock Model selection and estimation in regression with grouped variables.
\newblock {\em Journal of the Royal Statistical Society: Series B (Statistical
  Methodology)}, 68(1):49--67, 2006.

\bibitem{zhang2015}
Y.~Zhang and L.~Xiao.
\newblock Stochastic primal-dual coordinate method for regularized empirical
  risk minimization.
\newblock In {\em International Conference on Machine Learning (ICML)}, 2015.

\bibitem{zhu2015}
Z.~Zhu and A.~J. Storkey.
\newblock Adaptive stochastic primal-dual coordinate descent for separable
  saddle point problems.
\newblock In {\em Machine Learning and Knowledge Discovery in Databases}, pages
  645--658. Springer, 2015.

\end{thebibliography}

\end{document}